\documentclass[10pt,twocolumn,letterpaper]{article}

\usepackage{iccv}
\usepackage{times}
\usepackage{epsfig}
\usepackage{graphicx}
\usepackage{amsmath}
\usepackage{amssymb}

\usepackage{microtype}
\usepackage{subfigure}
\usepackage{multirow}
\usepackage{booktabs}
\usepackage{amsmath,amsfonts,bm}

\usepackage[pagebackref=true,breaklinks=true,letterpaper=true,colorlinks,bookmarks=false]{hyperref}
\newcommand{\ayse}[1]{{\color{black} #1} }
\newcommand{\ning}[1]{{\color{black} #1} }
\newcommand{\guilin}[1]{{\color{black} #1} }
\newcommand*{\affaddr}[1]{#1}
\newcommand*{\affmark}[1][*]{\textsuperscript{#1}}

\iccvfinalcopy 

\def\iccvPaperID{8068} 

\ificcvfinal\fi

\begin{document}

\title{Dual Contrastive Loss and Attention for GANs}

\author{Ning Yu\affmark[1,2]\hspace{0.5cm}
Guilin Liu\affmark[3]\hspace{0.5cm}
Aysegul Dundar\affmark[3,4]\\
Andrew Tao\affmark[3]\hspace{0.5cm}
Bryan Catanzaro\affmark[3]\hspace{0.5cm}
Larry Davis\affmark[1]\hspace{0.5cm}
Mario Fritz\affmark[5]\\
\affaddr{\affmark[1]University of Maryland}\hspace{0.5cm}
\affaddr{\affmark[2]Max Planck Institute for Informatics}\hspace{0.5cm}
\affaddr{\affmark[3]NVIDIA}\\
\affaddr{\affmark[4]Bilkent University}\hspace{0.5cm}
\affaddr{\affmark[5]CISPA Helmholtz Center for Information Security}\\
\tt\small \{ningyu,lsdavis\}@umd.edu\\
\tt\small \{guilinl,adundar,atao,bcatanzaro\}@nvidia.com\\
\tt\small fritz@cispa.saarland
}

\maketitle
\ificcvfinal\thispagestyle{empty}\fi

\begin{abstract}
Generative Adversarial Networks (GANs) produce impressive results on unconditional image generation when powered with large-scale image datasets. Yet generated images are still easy to spot especially on datasets with high variance (e.g. bedroom, church). In this paper, we propose various improvements to further push the boundaries in image generation. Specifically, we propose a novel dual contrastive loss and show that, with this loss, discriminator learns more generalized and distinguishable representations to incentivize generation. In addition, we revisit attention and extensively experiment with different attention blocks in the generator. We find attention to be still an important module for successful image generation even though it was not used in the recent state-of-the-art models. Lastly, we study different attention architectures in the discriminator, and propose a reference attention mechanism. By combining the strengths of these remedies, we improve the compelling state-of-the-art Fr\'{e}chet Inception Distance (FID) by at least 17.5\% on several benchmark datasets. We obtain even more significant improvements on compositional synthetic scenes (up to 47.5\% in FID). Code and models are available at \href{https://github.com/ningyu1991/AttentionDualContrastGAN}{GitHub}.

\end{abstract}


\section{Introduction}

\begin{figure}
\includegraphics[width=0.48\textwidth]{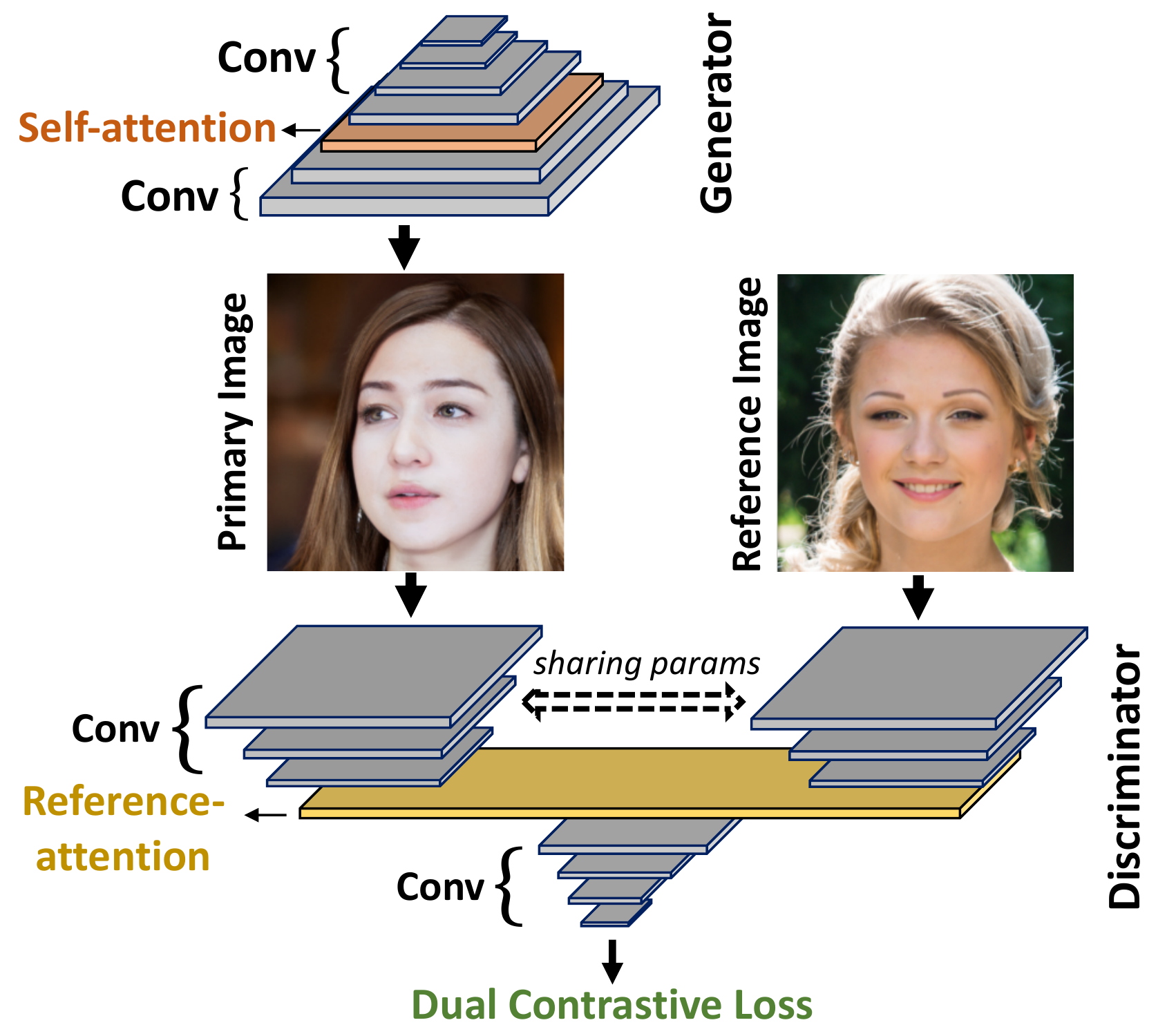}
\caption{The diagram of our GAN framework using three key components: self-attention in the generator, reference-attention in the discriminator, and a novel dual contrastive loss. Technical diagrams are in Fig.~\ref{fig:contrastive} and \ref{fig:attention_diagram}.}
\label{fig:teaser}
\end{figure}

Photorealistic image generation has increasingly become reality, benefiting from the invention of generative adversarial networks (GANs)~\cite{gan14nips} and its successive breakthroughs~\cite{radford2015dcgan,arjovsky2017wasserstein,gulrajani2017improved,miyato2018spectral,brock2018BigGAN,karras2017ProGAN,karras2019StyleGAN,karras2019StyleGAN2}. The progress is mainly driven by large-scale datasets~\cite{deng2009imagenet,liu2015deep,yu15lsun,johnson2017clevr,liu2017unsupervised,karras2019StyleGAN}, architectural tuning~\cite{chen2018self,zhang2019self,karras2019StyleGAN,karras2019StyleGAN2,schonfeld2020u}, and loss designs~\cite{mao2017least,arjovsky2017wasserstein,gulrajani2017improved,miyato2018spectral,jolicoeur2019relativistic,zhang2019consistency,zhao2020improved,yu2020inclusive,kang2020contragan,zhao2020image,jeong2021contrad}. GAN techniques have been popularized into extensive computer vision applications, including but not limited to image translation~\cite{isola2017image,zhu2017unpaired,zhu2017toward,liu2017unsupervised,huang2018multimodal,wang2018high,park2019semantic,dundar2020panoptic,park2020contrastive}, postprocessing~\cite{ledig2017photo,sonderby2016amortised,kupyn2018deblurgan,kupyn2019deblurgan,ulyanov2018deep,pan2020exploiting,zhang2016colorful}, image manipulation~\cite{choi2018stargan,choi2020stargan,shen2020interfacegan,abdal2020styleflow,bhattad2021view,wang2021hijack}, texture synthesis~\cite{yu2019texture,liu2020transposer,mardani2020neural}, image inpainting~\cite{iizuka2017globally,liu2018image,yu2018generative,yu2019free}, and text-to-image generation~\cite{reed2016generative,zhang2017stackgan,zhang2018stackgan++,tan2019text2scene}.

Yet, behind the seemingly saturated performance of the state-of-the-art StyleGAN2~\cite{karras2019StyleGAN2}, there still persists open issues of GANs that make generated images surprisingly obvious to spot~\cite{yu2019attributing,wang2020cnn,durall2020watch,he2021beyond}. Hence, it is still necessary to revisit the fundamental generation power when other concurrent deep learning techniques keep advancing and creating space for GAN improvements.

We investigate methods to improve GANs in two dimensions.
In the first dimension, we work on the loss function. As the discriminator aims to model the intractable real data distribution via a workaround of real/fake binary classification, a more effective discriminator can back-propagate more meaningful signals for the generator to compete against.
However, the feature representations of discriminators are often not generalized enough to incentivize the adversarially evolving generator and are prone to forgetting previous tasks~\cite{chen2019self} or previous data modes~\cite{srivastava2017veegan,li2018implicit}. This often leads to the generated samples with discontinued semantic structures~\cite{lin2019coco,zhang2019self} or the generated distribution with mode collapse~\cite{srivastava2017veegan,yu2020inclusive}. To mitigate this issue, we propose to synergize generative modeling with the advancements in contrastive learning~\cite{oord2018representation,chen2020simple}.
In this direction, for the first time, we replace the logistic loss of StyleGAN2 with a newly designed dual contrastive loss.

In the second dimension, we revisit the architecture of both generator and discriminator networks. 
Specifically, many GAN-based image generators rely on convolutional layers to encode features.
In such design, long-range dependencies across pixels (e.g., large-size semantically correlated layouts) can only be formulated with a deep stack of convolutional layers.
This, however, does not favor the stability of GAN training because of the challenge to coordinate multiple layers desirably. The minimax formulation and the alternating gradient ascent-descent in the GAN framework further exacerbate such instability.
To circumvent this issue, attention mechanisms that support long-range modeling across image regions are incorporated into GAN models~\cite{zhang2019self, brock2018BigGAN}.
After that, however, StyleGAN2 claimed the state of the art with a novel architectural design without any attention mechanisms. Therefore, it turns not clear whether attention still improves results, which of the popular attention mechanisms~\cite{jia2016dynamic,wu2020visual,wang2018non,zhao2020exploring}
improves the most, and in return of how many additional parameters.
To answer these questions, we extensively study the role of attention in the current state-of-the-art generator, and during this study improve the results significantly.

In the discriminator, we again explore the role of attention as shown in Fig.~\ref{fig:teaser}. We design a novel reference attention mechanism in the discriminator where we allow two irrelevant images as the inputs at the same time: one input is sampled from real data as a reference, and the other input is switched between a real sample and a generated sample. The two inputs are encoded through two Siamese branches~\cite{bromley1993signature,chopra2005learning,taigman2014deepface,zagoruyko2015learning} and fused by a reference-attention module. In this way, we achieve to guide real/fake classification under the attention of the real world. Contributions are summarized as follow: 



\begin{itemize}
  \setlength\itemsep{-0.3em}
  \vspace{-5pt}
     \item We propose a novel dual contrastive loss in adversarial training that generalizes representation to more effectively distinguish between real and fake, and further incentivize the image generation quality.
    \item  We investigate variants of the attention mechanism in GAN architecture to mitigate the local and stationary issues of convolutions.
    \item  We design a novel reference-attention discriminator architecture that substantially benefits limited-scale datasets.
    \item We conduct extensive experiments on large-scale datasets and their smaller subsets. We show that our improvements on the loss function and on the generator hold in both scenarios. On the other hand, we find discriminator to behave differently based on the number of available images, and the reference-attention-based discriminator to be only improving on limited-scale datasets.
    \item We redefine the state of the art by improving FID scores by at least 17.5\% on several large-scale benchmark datasets. 
    We also achieve more realistic generation on the CLEVR dataset~\cite{johnson2017clevr} which poses different challenges from the other datasets: compositional scenes with occlusions, shadows, reflections, and mirror surfaces. It comes with 47.5\% FID improvement.
\end{itemize}

\section{Related work}

\textbf{Generative adversarial networks (GANs).} Since the invention of GANs~\cite{gan14nips}, there have been rapid progress to achieve photorealistic image generation~\cite{radford2015dcgan,arjovsky2017wasserstein,gulrajani2017improved,gulrajani2017improved,miyato2018spectral,brock2018BigGAN,karras2017ProGAN,karras2019StyleGAN,karras2019StyleGAN2}. 
Significant improvements are obtained by careful architectural designs for generators~\cite{chen2018self,zhang2019self,karras2019StyleGAN,karras2019StyleGAN2,schonfeld2020u}, discriminators~\cite{wang2018high, liu2019learning} and new regularization techniques~\cite{mao2017least,arjovsky2017wasserstein,gulrajani2017improved,miyato2018spectral,zhang2019consistency,zhao2020improved,yu2020inclusive,kang2020contragan,zhao2020image,jeong2021contrad}.
Architectural evolution in generators started from a multi-layer perceptron (MLP)~\cite{gan14nips} and moved to deep convolutional neural networks (DCNN)~\cite{radford2015dcgan}, to models with residual blocks~\cite{miyato2018spectral}, and recently style-based~\cite{karras2019StyleGAN, karras2019StyleGAN2} and attention-based~\cite{zhang2019self,brock2018BigGAN} models.
Similarly, discriminators evolved from MLP to  DCNN~\cite{radford2015dcgan}, however,
their design has not been studied as aggressively.
In this paper, we propose changes in both generators and discriminators, and for the loss function.

\textbf{Contrastive learning.} Contrastive learning targets a transformation of inputs into an embedding where
associated signals are brought together, and they are distanced from the other samples in the dataset~\cite{hadsell2006dimensionality,tschannen2019mutual,chen2020simple,chen2020big}. The same intuition behind contrastive learning has also been the base of  Siamese networks~\cite{bromley1993signature,chopra2005learning,taigman2014deepface,zagoruyko2015learning}.
Contrastive learning is shown to be an effective tool for unsupervised learning~\cite{oord2018representation, he2020momentum, wu2018unsupervised}, conditional image synthesis~\cite{park2020contrastive,kang2020contragan,zhao2020image}, and domain adaptation~\cite{ge2020self}. In this work, we study its effectiveness \ning{when it is closely coupled with the adversarial training framework and replaces the conventional adversarial loss for unconditional image generation. It is orthogonal to ~\cite{kang2020contragan,zhao2020image,jeong2021contrad,lee2021infomax} where their contrastive losses serve only as an incremental auxiliary to the conventional adversarial loss, apply to the generator rather than the discriminator, and/or require expensive class annotations or augmentation for generation.}

\textbf{Attention models.} Attention models have dominated the language modeling~\cite{vaswani2017attention, wu2019pay, dai2019transformer, devlin2018bert, yang2019xlnet}, and became popular among various computer vision problems from image recognition~\cite{dai2017deformable, wang2017residual, hu2018gather, hu2018squeeze, zhao2018psanet, zhu2019deformable, hu2019local, wu2020visual} to image captioning~\cite{xu2015show, yang2016stacked, chen2017sca} to video prediction~\cite{jia2016dynamic, wang2018non}.
They are proposed in various forms: spatial attention that reweights the convolution activations~\cite{zhang2019self, wang2018non, chi2020non}, in different channels~\cite{wang2017residual, hu2018gather, hu2018squeeze}, or a combination of them~\cite{chen2017sca, woo2018cbam, fu2019dual}.
Attention models with their reweighting mechanisms provide a possibility for long-range modeling across distant image regions. 
As attention models outperform others in various computer vision tasks, researchers were quick to incorporate them into unconditional image generation~\cite{chen2018self, zhang2019self, parmar2018image, brock2018BigGAN}, semantic-based image generation~\cite{liu2019learning, tang2020local}, and text-guided image manipulation models~\cite{li2020lightweight, pavllo2020convolutional}.
Even though attention models have already benefited the image generation tasks, we believe the results can be further improved by empowering the state-of-the-art image synthesis models~\cite{karras2019StyleGAN2} (attention not involved) with the most recent achievements in the attention modules~\cite{zhao2020exploring}.
In addition, we design a novel reference-attention architecture for the discriminator and show a further boost on limited-scale datasets.

\begin{figure*}
\centering
\includegraphics[width=0.8\linewidth]{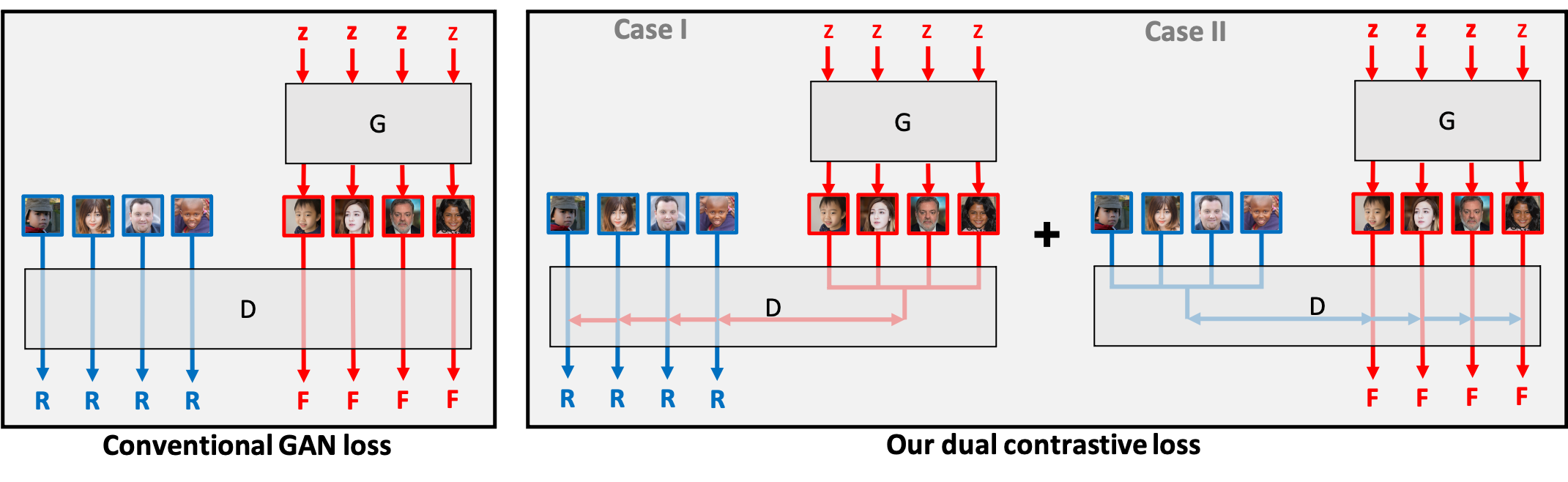}
\caption{Comparisons between the diagram of conventional GAN loss and diagram of our dual contrastive loss. Our contrastive loss in Case I  aims at teaching the discriminator to disassociate a single real image (R) against a batch of generated images (F). Dually in Case II, the discriminator learns to disassociate a single generated image against a batch of real images.}
\label{fig:contrastive}
\end{figure*}

\section{Approach}

Our improvements for GANs include a novel dual contrastive loss and variants of the attention mechanisms. For each improvement, we organize the context in a combination between method formulation and experimental investigation. After validating our optimal configuration, we compare it to the state of the art in Section~\ref{sec:exp}.

\subsection{Dual contrastive loss}
\label{sec:contrastive_loss}

Adversarial training relies on the discriminator's ability on real vs. fake classification. As in other classification tasks, discriminators are also prone to overfitting when the dataset size is limited \cite{arjovsky2017towards}.
On larger datasets, on the other hand, there is no study showing that disciminators overfit but we hypothesize that adversarial training can still benefit from novel loss functions which encourage the distinguishability power of the discriminator representations for their real vs. fake classification task.

We put another lens on the representation power of the discriminator by incentivizing generation via contrastive learning.
Contrastive learning associates data points and their positive examples and disassociates the other points within the dataset which are referred to as negative examples.
It is recently re-popularized by various unsupervised learning works~\cite{hadsell2006dimensionality,oord2018representation,tschannen2019mutual,chen2020simple,chen2020big} and generation works~\cite{park2020contrastive,kang2020contragan,zhao2020image}.
Among these works, contrastive learning is used as an auxiliary task.
For example in image to image translation task, a translator learns to output a zebra image given a horse image via adversarial loss and in addition learns to align the input horse image and the generated zebra image via contrastive loss function~\cite{park2020contrastive}.
Contrastive loss in that work is utilized such that given a patch showing the legs of an output zebra should be strongly associated with the corresponding legs of the input horse, more so than the other patches randomly extracted from the horse image.

In this work, different from the previous ones, we do not use contrastive learning as an auxiliary task but directly \ning{couple it in the main adversarial training} by a novel loss function formulation. We, to the best of our knowledge, for the first time train an unconditional GAN by solely relying on contrastive learning. As shown in Fig.~\ref{fig:contrastive} Right Case I, our contrastive loss function aims at teaching the discriminator to disassociate a single real image against a batch of generated images. Dually in Case II, the discriminator learns to disassociate a single generated image against a batch of real images.
The generator adversarially learns to minimize such dual contrasts.
Mathematically, we derive this loss function by extending the binary classification used in ~\cite{gan14nips,karras2019StyleGAN2} to a noise contrastive estimation framework~\cite{oord2018representation}, a one-against-a-batch classification in the softmax cross-entropy formulation. The novel formulation is as follows:

\noindent In Case I:
\begin{equation}
\begin{split}
L^\textit{contr}_\textit{real} & (G,D)=\underset{\mathbf{x}\sim p(\mathbf{x})}{\mathbb{E}}\left[\log \frac{e^{D(\mathbf{x})}}{e^{D(\mathbf{x})}+\underset{\mathbf{z}\sim \mathcal{N}(\mathbf{0},\mathbf{I}_d)}{\sum}e^{D(G(\mathbf{z}))}}\right]\\
& =-\underset{\mathbf{x}\sim p(\mathbf{x})}{\mathbb{E}}\left[\log\left(1+\sum_{\mathbf{z}\sim \mathcal{N}(\mathbf{0},\mathbf{I}_d)}e^{D(G(\mathbf{z}))-D(\mathbf{x})}\right)\right]\\
\end{split}
\label{eq:contrastive_real}
\end{equation}

\noindent In Case II:
\begin{equation}
\begin{split}
L^\textit{contr}_\textit{fake} & (G,D)=\underset{\mathbf{z}\sim \mathcal{N}(\mathbf{0},\mathbf{I}_d)}{\mathbb{E}}\left[\log \frac{e^{-D(G(\mathbf{z}))}}{e^{-D(G(\mathbf{z}))}+\underset{\mathbf{x}\sim p(\mathbf{x})}{\sum}e^{-D(\mathbf{x})}}\right]\\
& =-\underset{\mathbf{z}\sim \mathcal{N}(\mathbf{0},\mathbf{I}_d)}{\mathbb{E}}\left[\log\left(1+\sum_{\mathbf{x}\sim p(\mathbf{x})}e^{D(G(\mathbf{z}))-D(\mathbf{x})}\right)\right]\\
\end{split}
\label{eq:contrastive_fake}
\end{equation}

Comparing between Eq.~\ref{eq:contrastive_real} and \ref{eq:contrastive_fake}, the duality is formulated by switching the order of real/fake sampling while keeping the other calculation unchanged. Comparing to the logistic loss~\cite{gan14nips,karras2019StyleGAN2}, contrastive loss enriches the softplus formulation $\log(1+e^{D(\cdot)})$ with a batch of inner terms and using discriminator logit contrasts between real and fake samples.
Finally, our adversarial objective is:
\begin{equation}
\min_G\max_D\;L^\textit{contr}_\textit{real}(G,D)+L^\textit{contr}_\textit{fake}(G,D)
\label{eq:contrastive_adv_loss}
\end{equation}

\begin{table}[t]
\begin{center}
\resizebox{\linewidth}{!}{%
\begin{tabular}{l|ccccc}
\toprule
 & FFHQ & Bedroom & Church & Horse & CLEVR \\
\midrule
Non-saturating~\cite{gan14nips} (default) & 4.86 & 4.01 & 4.54 & 3.91 & 9.62 \\ 
Saturating~\cite{gan14nips} & 5.16 & 4.26 & 4.80 &5.90 & 10.46 \\ 
Wasserstein~\cite{gulrajani2017improved}  & 7.99 & 6.05 & 6.28 & 7.23 & \textbf{5.82} \\ 
Hinge~\cite{lim2017geometric} & 4.14 & 4.92 & 4.39 & 5.27 & 14.87 \\ 
Dual contrastive  (\textbf{ours}) & \textbf{3.98} &  \textbf{3.86} & \textbf{3.73}& \textbf{3.70} & 6.06 \\
\bottomrule
\end{tabular}}
\end{center}
\caption{Comparisons in FID among different GAN losses. Based on StyleGAN2 config E backbone, it shows our contrastive loss outperforms a variety of other losses on four out of five large-scale datasets. Wasserstein loss is better than ours on CLEVR, but are the worst on the other datasets.}
\label{tab:different_losses}
\end{table}

\begin{table}[]
\begin{center}
\small
\resizebox{\linewidth}{!}{
\begin{tabular}{l|ccccc}
\midrule
Loss & FFHQ & Bedroom & Church & Horse & CLEVR \\
\midrule
Non-saturating~\cite{gan14nips} (default) & 245. & 332. & 517. & 1285. & 199. \\
Dual contrastive  (\textbf{ours}) & \textbf{377.} & \textbf{580.} & \textbf{856.} & \textbf{1645.} & \textbf{513.} \\
\bottomrule
\end{tabular}}
\end{center}
\caption{Comparisons in FDDF between StyleGAN2 default loss and our loss. A larger value is more desirable, indicating the learned discriminator features are more distinguishable between real and fake.}
\label{tab:contrast_ablation}
\end{table}

\begin{figure}
\centering
\includegraphics[width=\linewidth]{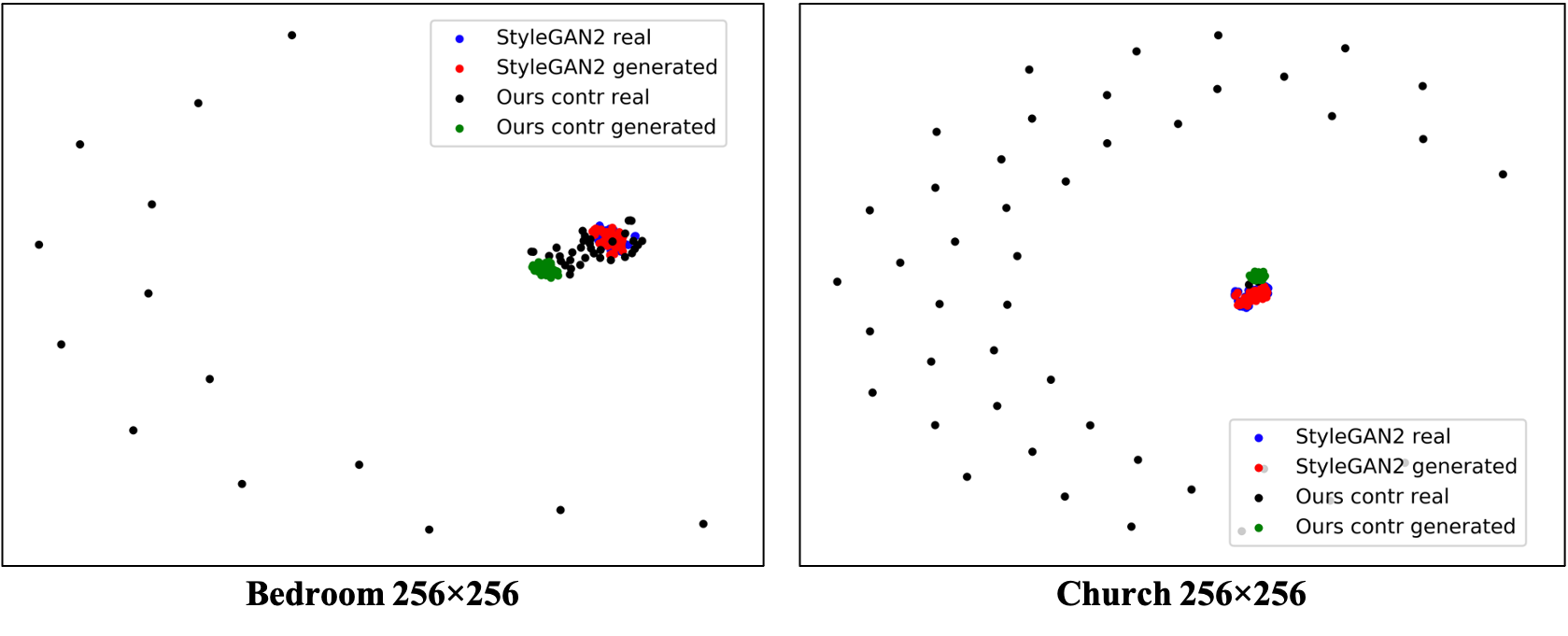}
\caption{The tSNE plots for the distributions of discriminator features. The distinguishability of features based on our contrastive loss is much more significant than that based on the default non-saturating loss in StyleGAN2 baseline. Our loss learns to associate fake features to a ``core'' clique (green) while pushing real features in the wild outwards as ``satellites'' (black). The baseline loss fails to differentiate features from the two sources (red v.s. blue) with a clear margin.}
\label{fig:contrastive_tsne}
\end{figure}

\textbf{Investigation on loss designs.}
We extensively validate the effectiveness of dual contrastive loss compared to other loss functions as presented in Table~\ref{tab:different_losses}.
We replace the loss used in StyleGAN2~\cite{karras2019StyleGAN2}, non-saturating default loss, with other popular GAN losses while keeping all the other parameters the same.
As shown in Table~\ref{tab:different_losses}, dual contrastive loss is the only loss that significantly improves upon the default loss of StyleGAN2 consistently on all the five datasets. Wasserstein loss is better than ours on CLEVR dataset, but is the worst among all the loss functions on the other datasets. We reason the success of the dual loss to its formulation that explicitly learns an unbiased representation between real and generated distributions.



\textbf{The distinguishability of contrastive representation.}
Motivated by the consistent improvement from our dual contrastive loss, we delve deeper to investigate if and by how much our contrastive representation is more distinguishable than the original discriminator representation.
We measure the representation distinguishability by the Fr\'{e}chet distance of the discriminator features in the last layer (FDDF) between 50K real and generated samples.
A larger value indicates more distinguishable features  between real and fake. 
We find our dual contrastive features to be consistently more distinguishable than the original discriminator features as shown in Table~\ref{tab:contrast_ablation} and Fig.~\ref{fig:contrastive_tsne}, which back-propagates more effective gradients to incentivize our generator.



\begin{figure*}
\centering
\includegraphics[width=0.94\textwidth]{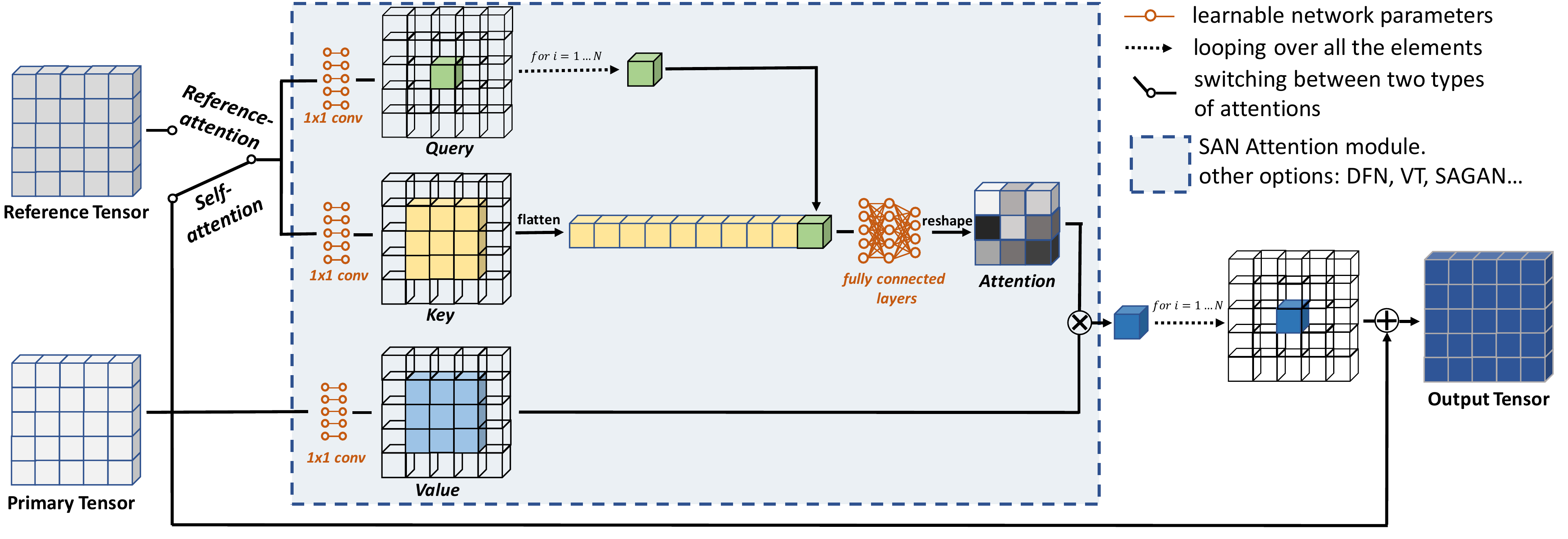}
\caption{The diagram of self-attention and reference-attention schemes. \ning{The attention module is instantiated by SAN~\cite{zhao2020exploring} but is agnostic to network backbone. It can flexibly switch to other options and be plug-and-play.} We switch between the sources that are used to calculate the Key and Query tensors, so as to implement self-attention and reference-attention respectively.}
\label{fig:attention_diagram}
\end{figure*}

\subsection{Self-attention in the generator}
\label{sec:self_attn_G}

The majority of the GAN-based image generators rely solely on convolutional layers to extract features~\cite{radford2015dcgan,arjovsky2017wasserstein,gulrajani2017improved,miyato2018spectral,karras2017ProGAN,karras2019StyleGAN,karras2019StyleGAN2}, even though the local and stationary convolution primitive in the generator can not model the long-range dependencies in an image. Among recent GAN-based models, SAGAN~\cite{zhang2019self} uses the self-attention block~\cite{wang2018non} and demonstrates improved results. BigGAN~\cite{brock2018BigGAN} also follows this choice and uses a similar attention module for better performance.
After that, however, StyleGAN~\cite{karras2019StyleGAN} and StyleGAN2~\cite{karras2019StyleGAN2} redefine the state of the art with various modifications in the generator architecture which do not include any attention mechanisms. StyleGAN2 also shows that generation results can be improved by larger networks with an increased number of convolution filters. Therefore, it is now not clear what the role of attention is in the state-of-the-art image generation models.
Does attention still improve the network performance? Which attention mechanism benefits the most and in the trade of how many additional parameters?
To answer these questions, we experiment with previously proposed self-attention modules: Dynamic Filter Networks (DFN)~\cite{jia2016dynamic}, Visual Transformers (VT)~\cite{wu2020visual}, Self-Attention GANs (SAGAN)~\cite{zhang2019self}, as well as the state-of-the-art patch-based spatially-adaptive self-attention module, SAN~\cite{zhao2020exploring}.

\ning{All the above self-attention modules are benefited from their adaptive data-dependent parameter space while they have their own hand-crafted architecture designs and interpretability. DFN~\cite{jia2016dynamic} keeps the convolution primitive but makes the convolutional filter condition to its input tensor. VT~\cite{wu2020visual} compresses input tensor to a set of 1D feature vectors, interprets them as semantic tokens, and leverages language transformer~\cite{vaswani2017attention} for tensor propagation. SAN~\cite{zhao2020exploring} generalizes the self-attention block~\cite{wang2018non} (as used in SAGAN~\cite{zhang2019self}) by replacing the point-wise softmax attention with a patch-wise fully-connected transformation.

We show the diagram of self-attention in Figure~\ref{fig:attention_diagram}, with a specific instantiation from SAN~\cite{zhao2020exploring} due to its generalized and state-of-the-art design.
Note that the attention module is agnostic to network backbone and can be switched to other options for fair comparisons. For conceptual and technical completeness, we formulate our SAN-based self-attention below.} 

In details, let $\mathbf{T}\in\mathbb{R}^{h \times w \times c}$ be the input tensor to a convolutional layer in the original architecture. Following the mainstream protocol of self-attention calculation~\cite{wang2018non,zhang2019self,parmar2018image}, we obtain the corresponding key, query, and value tensors \ayse{$\mathbf{K(T)},\mathbf{Q(T)},\mathbf{V(T)}\in\mathbb{R}^{h \times w \times c}$ separately using $1\times1$ convolutional kernel followed by bias and leaky ReLU.} 
For each location $(i,j)$ within the tensor spatial dimensions, we extract a large patch with size $s$ from $\mathbf{K}$ centered at $(i,j)$, denoted as $\mathbf{k}\in\mathbb{R}^{s \times s \times c}$. We then flatten the patch and concatenate it along the channel dimension with $\mathbf{q}\in\mathbb{R}^{1 \times 1 \times c}$, the query vector at $(i,j)$, to obtain $\mathbf{p}\in\mathbb{R}^{1 \times 1 \times (s^2c+c)}$:
\begin{equation}
\begin{split}
\mathbf{k} & =\mathbf{K}_{(i-\frac{s}{2}:i+\frac{s}{2}+1,\;j-\frac{s}{2}:j+\frac{s}{2}+1)}\\
\mathbf{q} & =\mathbf{Q}_{(i,j)}\\
\mathbf{p} & =\mathsf{concat}\left(\mathsf{flatten}(\mathbf{k}),\mathbf{q}\right)\\
\end{split}
\label{eq:concatenate}
\end{equation}

In order to cooperate between the key and query, we feed $\mathbf{p}$ through two fully-connected layers followed by bias and leaky ReLU and obtain a vector with  size $\tilde{\mathbf{w}}\in\mathbb{R}^{1 \times 1 \times s^2c}$:
\begin{equation}
\begin{split}
\hat{\mathbf{w}} & =\mathsf{leakyReLU}(\mathbf{p}\mathbf{M}_{w1}+\mathbf{b}_{w1})\\
\tilde{\mathbf{w}} & =\hat{\mathbf{w}}\mathbf{M}_{w2}+\mathbf{b}_{w2}\\
\end{split}
\label{eq:fc}
\end{equation}
$\mathbf{M}_{w1}\in\mathbb{R}^{(s^2c+c) \times s^2c}$, $\mathbf{M}_{w2}\in\mathbb{R}^{s^2c \times s^2c}$, and $\mathbf{b}_{w1},\mathbf{b}_{w2}\in\mathbb{R}^{1 \times 1 \times s^2c}$ are the learnable parameters in the fully connected layers and biases.

On one hand we reshape $\tilde{\mathbf{w}}$ back to the patch size $\mathbf{w}\in\mathbb{R}^{s \times s \times c}$; on the other hand we extract a patch with the same size from $\mathbf{V}$ centered at $(i,j)$, denoted as $\mathbf{v}\in\mathbb{R}^{s \times s \times c}$. Next, we aggregate $\mathbf{v}$ over spatial dimensions with the correponding weights from $\mathbf{w}$ to derive an output vector $\mathbf{o}\in\mathbb{R}^{1 \times 1 \times c}$:
\begin{equation}
\begin{split}
\mathbf{w} & =\mathsf{reshape}(\tilde{\mathbf{w}})\\
\mathbf{v} & =\mathbf{V}_{(i-\frac{s}{2}:i+\frac{s}{2}+1,\;j-\frac{s}{2}:j+\frac{s}{2}+1)}\\
\mathbf{o}(i,j) & =\sum_{m,n=1}^s \mathbf{w}_{(m,n)}\mathbf{v}_{(m,n)}\\
\end{split}
\label{eq:output}
\end{equation}

We loop over all the $(i,j)$ to constitute an output tensor $\bar{\mathbf{O}}^\textit{self}\in\mathbb{R}^{h \times w \times c}$ and define it as the self-attention output. Finally, we replace the original convolution output with $\mathbf{O}^\textit{self}\in\mathbb{R}^{h \times w \times c}$, a residual version of this self-attention output.
\begin{equation}
\begin{split}
\bar{\mathbf{O}}^\textit{self}_{(i,j)} & =\mathbf{o}(i,j),\;\;\forall i=1,\dots,h,\;j=1,\dots,w\\
\bar{\mathbf{O}}^\textit{self} & \doteq\mathsf{attn}\left(\mathbf{K}(\mathbf{T}),\mathbf{Q}(\mathbf{T}),\mathbf{V}(\mathbf{T})\right)\\
\mathbf{O}^\textit{self} & =\bar{\mathbf{O}}^\textit{self}+\mathbf{T}\\
\end{split}
\label{eq:output_self_attn}
\end{equation}

\begin{table}[]
\begin{center}
\small
\resizebox{\linewidth}{!}{
\begin{tabular}{l|cccc}
\midrule
 & CelebA & Animal Face & Bedroom & Church  \\
\midrule
StyleGAN2~\cite{karras2019StyleGAN2}  & 9.84  & 36.55  & 19.33 &  11.02 \\
+ DFN~\cite{jia2016dynamic}  & \textbf{8.41}  & 35.10 & 26.86  & 11.31 \\
+ VT~\cite{wu2020visual} & 9.18  & 34.70  & 16.85  & 10.64  \\
+ SAGAN~\cite{zhang2019self} & 9.35 & 34.83  & 17.94  & 10.65  \\
+ SAN~\cite{zhao2020exploring}& 8.60  & \textbf{32.72} & \textbf{16.36}  & \textbf{9.62}  \\
\bottomrule
\end{tabular}}
\end{center}
\caption{Comparisons in FID among different attention modules in the generator. StyleGAN2 config E which does not include an attention module is used as a backbone. For computationally efficient comparisons, we use the 30k subset of each dataset at 128$\times$128 resolution.}
\label{tab:attn_ablation}
\end{table}

\begin{figure}
\centering
\includegraphics[width=\linewidth]{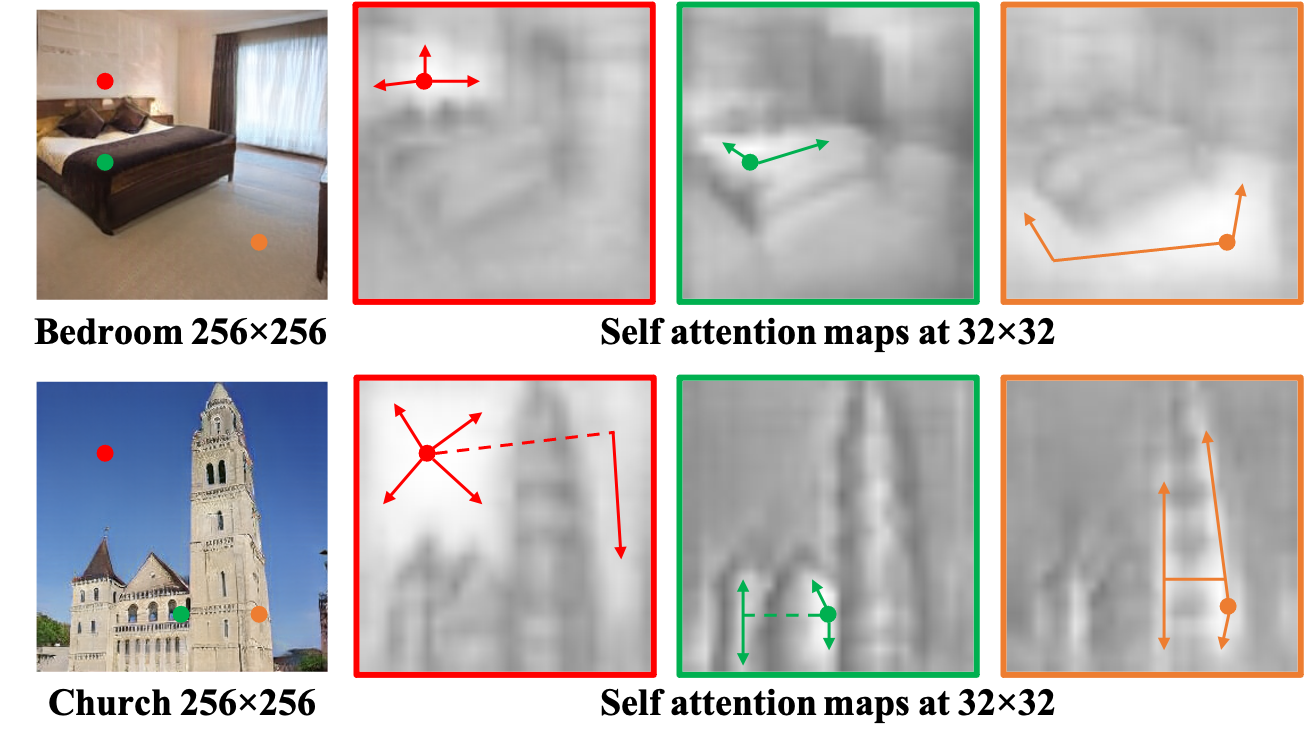}
\caption{StyleGAN2 + SAN generated samples and their self-attention maps in the generator for the corresponding dot positions. Considering there is an attention weight kernel $\mathbf{w}\in\mathbb{R}^{s \times s \times c}$ for each position, we visualize the norm for each spatial position of $\mathbf{w}$. The attention maps strongly align to the semantic layout and structures of the generated images, which enable long-range dependencies across objects. \ning{See more samples in the supplementary material.}
}
\label{fig:attention}
\end{figure}

It is worth noting that $\mathbf{w}$ plays a conceptually equivalent role as the softmax attention map of the traditional key-query aggregation~\cite{wang2018non,zhang2019self,parmar2018image}, except it is not identical across channels anymore but rather generalized to optimize for each channel. $\mathbf{w}$ also aligns in spirit with the concept of DFN~\cite{jia2016dynamic}, except the spatial size $s$$\times$$s$ is empirically set much larger than 3$\times$3, and more importantly, $\mathbf{w}$ is not ``sliding'' anymore but rather generalized to optimize at each location.

\ning{\textbf{Investigations on self-attention modules.}
In Table~\ref{tab:attn_ablation} we extensively compare among a variety of self-attention modules by replacing the default convolution in the 32$\times$32-resolution layer in StyleGAN2~\cite{karras2019StyleGAN2} config E backbone with one of them. We justify that SAN~\cite{zhao2020exploring} significantly improves over the StyleGAN2 baseline and outperforms the other attention variants on several datasets. DFN~\cite{jia2016dynamic} is better than ours on CelebA dataset, but is the worst on most other datasets. We provide additional ablation studies on network architectures in the supplementary material. }

We visualize the attention map examples of the best performing generator (StyleGAN2 + SAN) in Fig. \ref{fig:attention}. We find attention maps to strongly correspond to the semantic layout and structures of the generated images.

\ning{\textbf{Complexity of self-attention modules.}
We also compare in Table~\ref{tab:attn_complexity} the time and space complexity of these self-attention modules. We observe that DFN~\cite{jia2016dynamic} and VT~\cite{wu2020visual} moderately improve the generation quality yet in the trade of undesirable $>3.6\times$ complexity. On the contrary, the improvements from SAGAN~\cite{zhang2019self} or SAN~\cite{zhao2020exploring} are not at the cost of complexity, but rather benefited from the more representative attention designs. They use a fewer number of convolution channels and the multi-head trick~\cite{wang2018non} to control their complexity.}
These results show that the improved performance does not come from any additional parameters but rather the attention structure itself.

\begin{table}[t!]
\begin{center}
\small
\begin{tabular}{l|cc}
\toprule
Method & FLOPS (G) & \#parameters (M)  \\
\midrule
StyleGAN2~\cite{karras2019StyleGAN2} & 1.08 & 48.77 \\
+ DFN~\cite{jia2016dynamic} & 4.20 & 177.60 \\
+ VT~\cite{wu2020visual} & 7.39 & 240.09\\
+ SAGAN~\cite{zhang2019self} & 0.99 & 44.99\\
+ SAN~\cite{zhao2020exploring}& 1.08 & 48.43 \\
\bottomrule
\end{tabular}
\end{center}
\caption{Time complexity in FLOPS and space complexity in the number of parameters for each method.}
\label{tab:attn_complexity}
\end{table}
\begin{figure*}[t!]
\center
  \subfigure{
    \centering
    \includegraphics[width=0.23\linewidth]{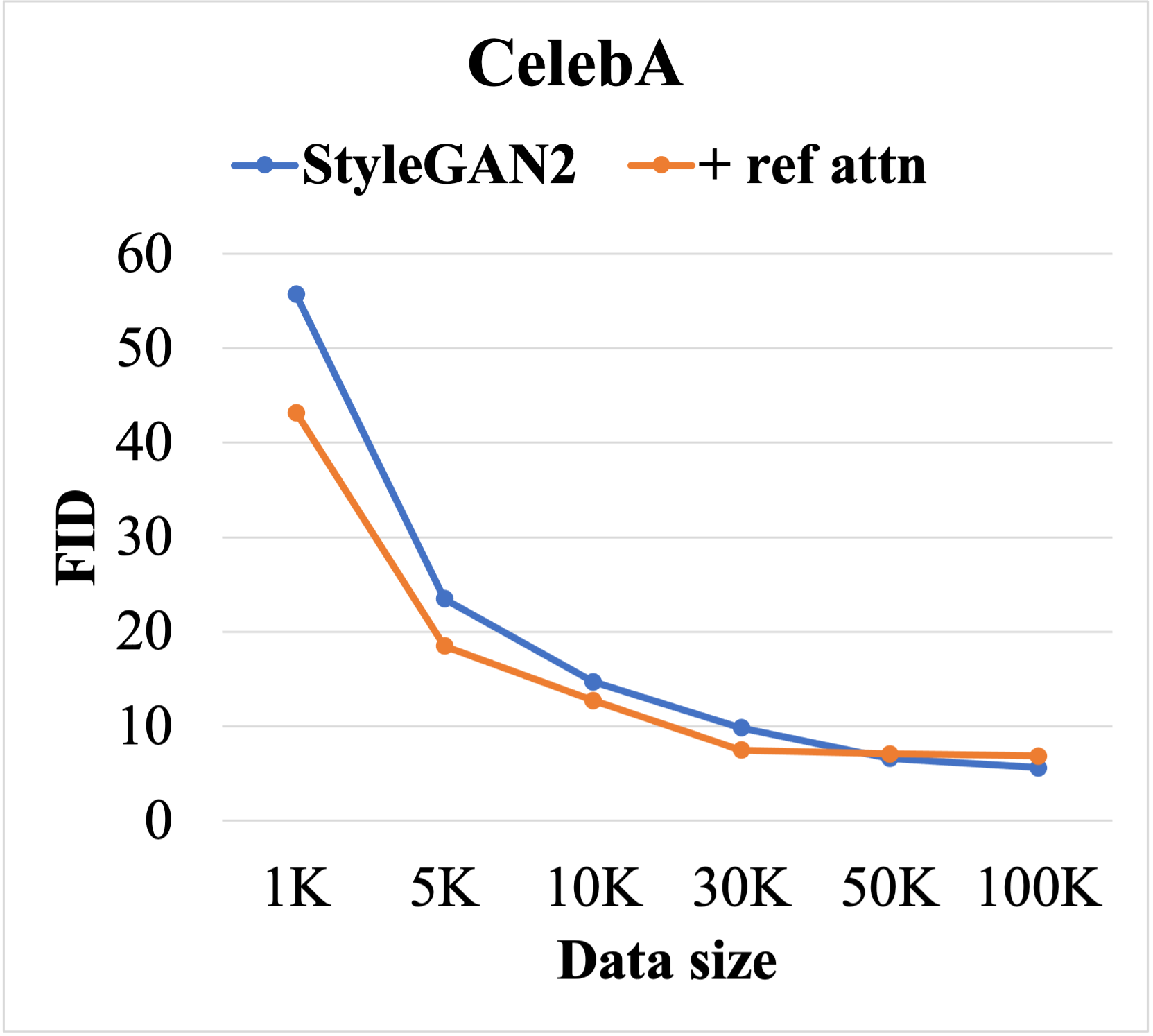}
  }
  \subfigure{
    \centering
    \includegraphics[width=0.23\linewidth]{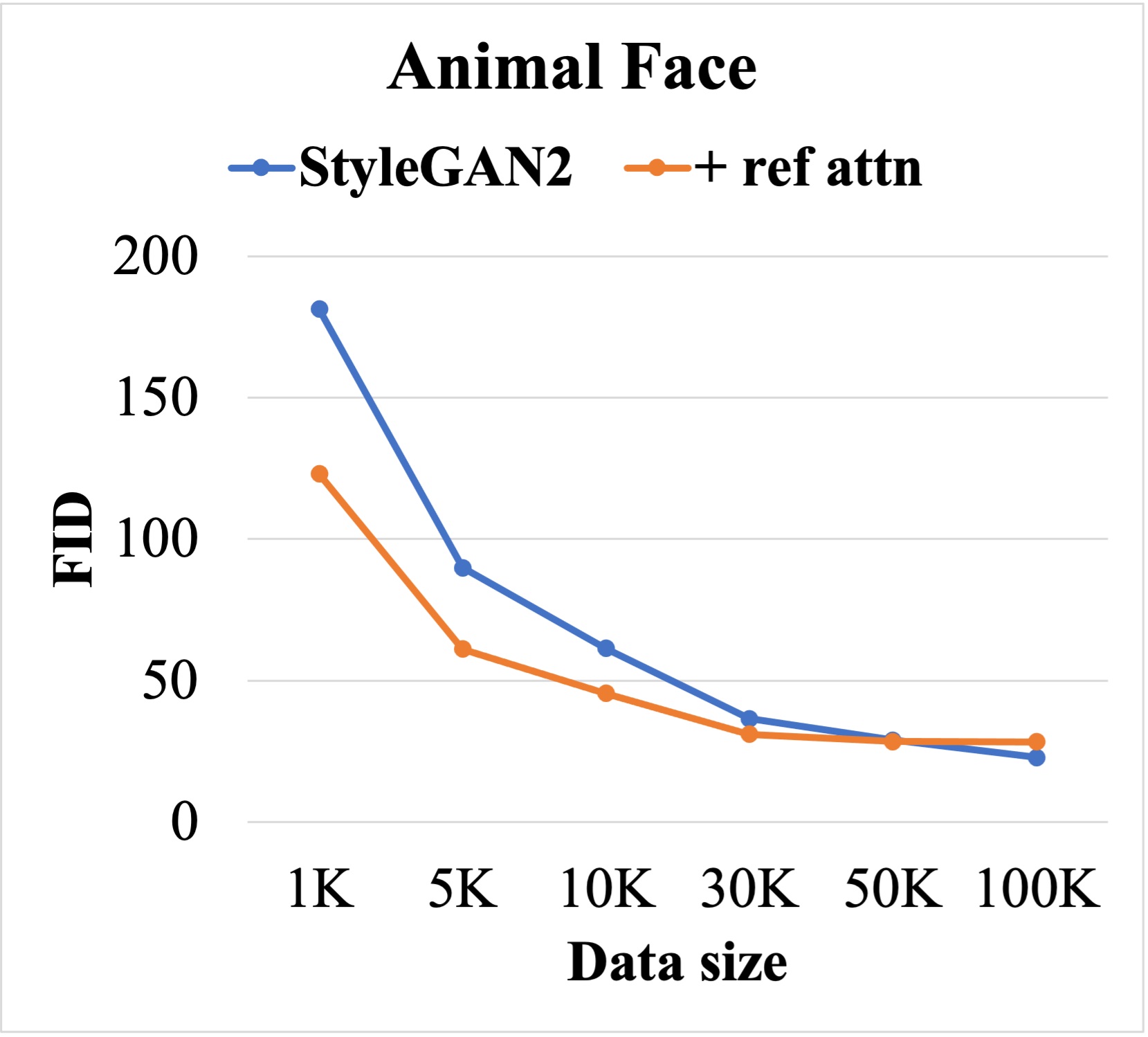} 
  }
  \subfigure{
    \centering
    \includegraphics[width=0.23\linewidth]{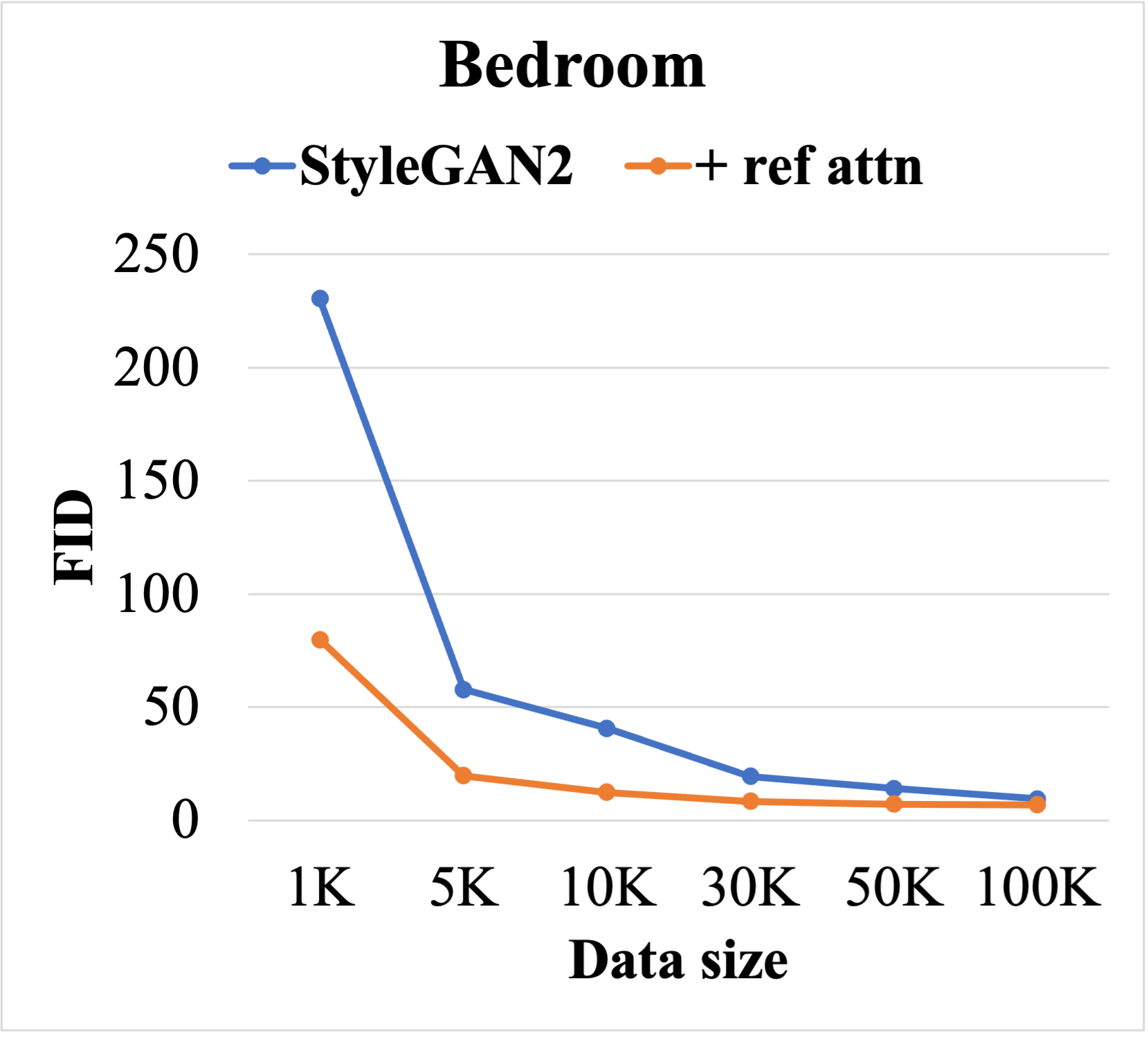} 
  }
  \subfigure{
    \centering
    \includegraphics[width=0.23\linewidth]{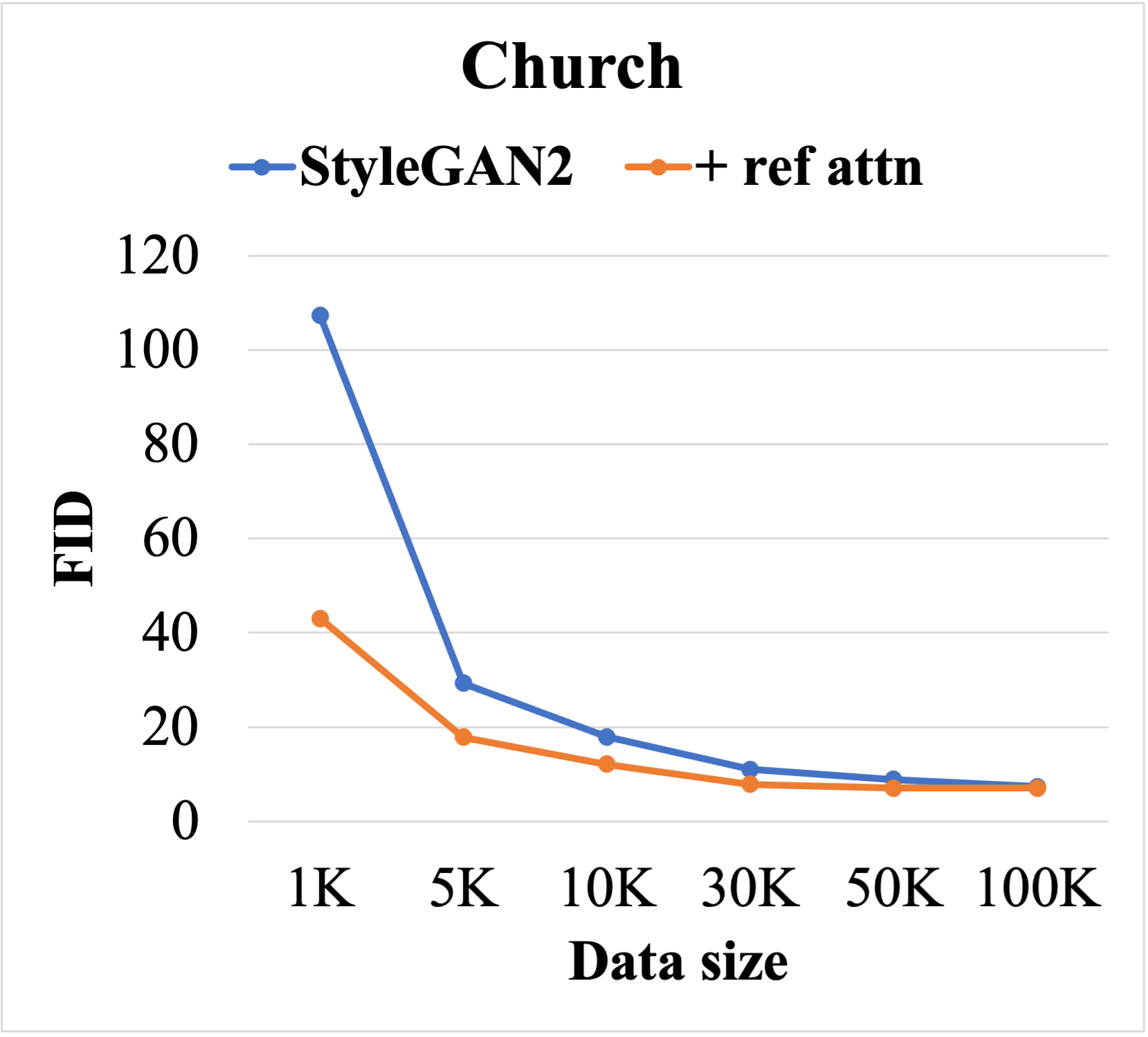} 
  }
  \caption{Comparisons in FID between StyleGAN2 config E baseline (blue) and that with our reference-attention in the discriminator (orange). Our method consistently improves the baseline when dataset size varies between 1k and 30k images. For computationally efficient comparisons, we use each dataset at 128$\times$128 resolution. See the supplementary material for the values in these plots.}
  \label{fig:comparisons_dataset_size} 
\end{figure*}

\subsection{Reference-attention in the discriminator}
\label{sec:ref_attn_D}

First, we apply SAN~\cite{zhao2020exploring}, the best attention mechanism we validated in the generator, to the discriminator. However, we do not see a benefit of such design as shown in Table \ref{tab:analysis_D_res}.
Then, we explore an advanced attention scheme given that two classes of input (real vs. fake) are fed to the discriminator.
We allow the discriminator to take two image inputs at the same time: the \textit{reference} image and the \textit{primary} image where we set the reference image to always be a real sample while the primary image to be either a real or generated sample.
The reference image is encoded to represent one part of the attention components. These components are learned to guide the other part of the attention components, which are encoded from the \textit{primary} image. There are three insights in this advancement. (1) An effective discriminator encodes real images and generated images differently, so that reference-attention is capable of learning positive feedback given both images from the real class and negative feedback given two images from different classes. Such a scheme amplifies the representation difference between real and fake, and in turn potentially strengthens the power of the discriminator. (2) Reference-attention enables distribution estimation in the discriminator feature level beyond the discriminator logit level in the original GAN framework, which guides generation more strictly towards the real distribution. (3) Reference-attention learns to cooperate real and generated images explicitly in one round of back-propagation, instead of individually classifying them and trivially averaging the gradients over one batch. Arbitrarily pairing up images mitigates discriminator from overfitting, similar to the spirit of random data augmentation, but we instead conduct random feature augmentation using attention.

In detail, we first encode the reference image and the primary image through the original discriminator layers prior to the convolution at a certain resolution. To align feature embeddings, we apply the Siamese architecture~\cite{bromley1993signature,chopra2005learning} to share layer parameters as shown in Fig.~\ref{fig:teaser}.
We then apply the same attention scheme as used in the generator, except we use the tensor $\mathbf{T}_\textit{ref}\in\mathbb{R}^{h \times w \times c}$ from the reference branch to calculate the key and query tensors, and use the tensor $\mathbf{T}_\textit{pri}\in\mathbb{R}^{h \times w \times c}$ from the primary branch to calculate the value tensor and the residual shortcut. Finally, we replace the original convolution output with our reference-attention output:
\begin{equation}
\mathbf{O}^\textit{ref}\doteq\mathsf{attn}\left(\mathbf{K}(\mathbf{T}_\textit{ref}),\mathbf{Q}(\mathbf{T}_\textit{ref}),\mathbf{V}(\mathbf{T}_\textit{pri})\right)+\mathbf{T}_\textit{pri}
\label{eq:output_ref_attn}
\end{equation}

After the reference-attention layer, the two Siamese branches fuse into one and are followed by the remaining discriminator layers to obtain the classification logit. We show in Fig.~\ref{fig:attention_diagram} the diagram of reference-attention. 
Eq.~\ref{eq:output_ref_attn} provides the flexibility how to cooperate between reference and primary images. \ning{We empirically explore the other compositions of sources to the key, query, and value components of reference-attention in the supplementary material as well as additional ablation studies on network architectures.}


\begin{table}[t!]
\begin{center}
\small
\resizebox{\linewidth}{!}{
\begin{tabular}{l|cccc}
\toprule
 & CelebA & Animal Face  & Bedroom  & Church  \\
\midrule
StyleGAN2~\cite{karras2019StyleGAN2} & 9.84 & 36.55 & 19.33 & 11.02 \\
+ self attn in D & 10.49 & 42.41 & 17.22 & 11.06 \\
+ ref attn in D & \textbf{7.48} & \textbf{31.08} & \textbf{8.32}& \textbf{7.86} \\
\bottomrule
\end{tabular}}
\end{center}
\caption{Comparisons in FID among different attention configurations in the discriminator. StyleGAN2 config E which does not include any attention module is used as a backbone. For computationally efficient comparisons, we use the 30k subset of each dataset at 128$\times$128 resolution.}
\label{tab:analysis_D_res}
\end{table}

From Table~\ref{tab:analysis_D_res} we validate reference-attention mechanism (ref attn) to improve the results whereas self-attention to be barely benefiting for the discriminator.
Encouraged with these findings, we run the proposed reference-attention on full-scale datasets but do not see any improvements. Therefore, we dive deep into reference-attentions behavior in the discriminator with respect to the dataset size as given in Fig.~\ref{fig:comparisons_dataset_size}.
We find that the reference-attention in the discriminator consistently improves the performance when dataset size varies between 1k and 30k images, and on contrary slightly deteriorates the performance when dataset sizes increase further.
\ayse{We reason that the arbitrary pair-up of the reference and primary image inputs to prevent overfitting when data size is small but causing underfitting with the increase of data size }
\ayse{Even though in this paper our main scope is GANs on large-scale datasets, we believe these findings to be very interesting for researchers to design their networks for limited-scale datasets.} We summarize our comparisons on limited-scale datasets in the supplementary material.


\begin{figure*}
\centering
\includegraphics[width=\linewidth]{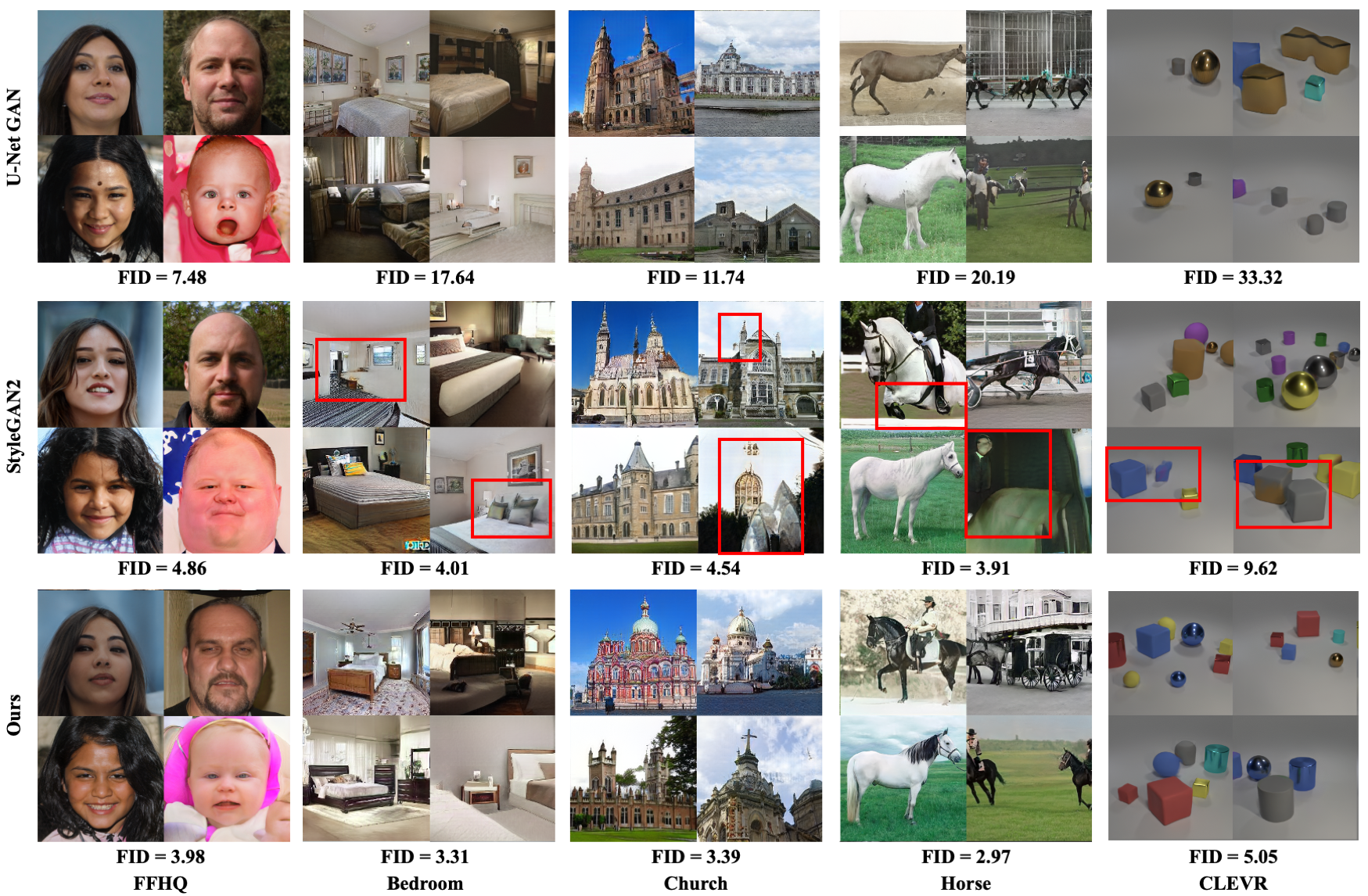}
\caption{Uncurated generated samples. To align the comparisons, we use the same real query images for pre-trained generators to reconstruct. Artifacts from StyleGAN2 are highlighted with red boxes. Zoom in for details. In particular, our generation significantly outperforms the baselines on CLEVR images which strongly rely on long-range dependencies (occlusions, shadows, reflections, etc) and consistency (consistent shadow directions, consistent specularity, regular shapes, uniform colors, etc). \ning{See more samples in the supplementary material.}}
\label{fig:samples}
\end{figure*}

\section{Comparisons to the state of the art}
\label{sec:exp}



\textbf{Implementation details.} All our models are built upon the most recent state-of-the-art unconditional StyleGAN2~\cite{karras2019StyleGAN2} config E for its high performance and reasonable speed.
We leverage the plug-and-play advantages of all our improvement proposals to strictly follow StyleGAN2 official setup and training protocol, which facilitates reproducibility and fair comparisons. For dual contrastive loss, we first warm up training with the default non-saturating loss for about 20 epochs, and then switch to train with our loss.

\textbf{Datasets.} We use several benchmark datasets, 70K FFHQ face dataset~\cite{karras2019StyleGAN}, 3M LSUN Bedroom dataset~\cite{yu15lsun}, 120K LSUN Church dataset~\cite{yu15lsun}, 2M LSUN Horse dataset~\cite{yu15lsun}, CelebA face dataset~\cite{liu2015deep} and  Animal Face dataset~\cite{liu2019few}, and 70K CLEVR~\cite{johnson2017clevr} dataset which contains rendered images with random compositions of 3D shapes, uniform materials, uniform colors,  point lighting, and a plain background.
It poses different challenges from the other common datasets: compositional scenes with occlusions, shadows, reflections, and mirror surfaces.
We use 256$\times$256 resolution images for each of these datasets except the CelebA and Animal Face datasets which are used in 128$\times$128 resolutions.
We do not experiment with 1024$\times$1024 resolution of FFHQ as it takes 9 days to train StyleGAN2 base model.
Instead, we run extensive experiments on the mentioned various datasets.
If not otherwise noted, we use the whole dataset.


\textbf{Evaluation.} FID~\cite{heusel2017gans} is regarded as the golden standard to quantitatively evaluate generation quality. We follow the protocol in StyleGAN2~\cite{karras2019StyleGAN2} to report the FID between 50K generated images and 50K real testing images. \ning{The smaller the more desirable.}
In the supplementary material, we report various other metrics that are proposed in StyleGAN~\cite{karras2019StyleGAN} or StyleGAN2~\cite{karras2019StyleGAN2} but are less benchmarked in other literature, Perceptual Path Length, Precision,  Recall, and Separability.

\textbf{Comparisons.}
\label{sec:comparisons}
Besides StyleGAN2~\cite{karras2019StyleGAN2}, we also compare to a parallel state-of-the-art study, U-Net GAN~\cite{schonfeld2020u}, which was build upon and improved on BigGAN~\cite{brock2018BigGAN}. We train U-Net by adapting it to the better backbone of  StyleGAN2~\cite{karras2019StyleGAN2} for fair comparison, and obtain better results than their official release on non-FFHQ datasets.
As shown in Table~\ref{tab:main_results}, our self-attention generator improves on four out of five large-scale datasets, up to 13.3\% relative improvement on Bedroom dataset. This highlights the benefits of attention to details and to long-range dependencies on complex scenes. 
However, self-attention does not improve on the extensively studied FFHQ dataset. We reason that the image pre-processing of facial landmark alignment compensates for the lack of attention schemes, which makes previous works also overlook them on other datasets.

\begin{table}[!t]
\begin{center}
\resizebox{\linewidth}{!}{%
\begin{tabular}{ll|ccccc}
\toprule
Method & Loss & FFHQ & Bedroom & Church & Horse & CLEVR \\
\midrule
BigGAN~\cite{brock2018BigGAN} &  Adv & 11.4 & - & - & - & - \\
U-Net GAN~\cite{schonfeld2020u} & Adv & 7.48 & 17.6 & 11.7 & 20.2 & 33.3 \\
StyleGAN2~\cite{karras2019StyleGAN2} & Adv & 4.86 & 4.01 & 4.54 & 3.91 & 9.62 \\
\midrule
StyleGAN2 w/ attn & Adv & 5.13 & \underline{3.48} & 4.38 & \underline{3.59} & 8.96 \\
StyleGAN2 & Contr  & \textbf{3.98} & 3.86 & \underline{3.73} & 3.70 & \underline{6.06} \\
StyleGAN2 w/ attn & Contr & \underline{4.63} & \textbf{3.31} & \textbf{3.39} & \textbf{2.97} & \textbf{5.05} \\
\bottomrule
\end{tabular}}
\end{center}
\caption{Comparisons in FID to the state-of-the-art GANs on the large-scale datasets. \ning{We highlight the best in \textbf{bold} and second best with \underline{underline}.} \guilin{``w/ attn'' indicates using the self-attention in the generator.} \ning{``Contr'' indicates using our dual contrastive loss instead of conventional GAN loss.}}
\label{tab:main_results}
\end{table}

Our dual contrastive loss improves effectively on all the datasets, up to 37\% improvement on CLEVR dataset. This highlights the benefits of contrastive learning on generalized representation, especially on aligned datasets, e.g. FFHQ and CLEVR, that can easily make a traditional discriminator overfit. The synergy effective between self-attention and contrastive learning is significant and consistent, resulting in at least 17.5\% and up to 47.5\% relative improvement on CLEVR.
Especially for CLEVR, our generator handles more realistically for occlusions, shadows, reflections, and mirror surfaces.
As shown in Fig. \ref{fig:samples}, our method suppresses artifacts that were previously visible in StyleGAN2 baseline outputs,  with red boxes, e.g., the artifacts on the wall in Bedroom images, discontinuities in the structure in Church images, as well as color leakage between objects in CLEVR images.




\section{Conclusion}

The advancements in attention schemes and contrastive learning generate opportunities for new designs of GANs. Our attention schemes serve as a beneficial replacement for local and stationary convolutions, so as to equip generation and discriminator representation with long-range adaptive dependencies. In particular, our reference-attention discriminator cooperates between real reference images and primary images, mitigates discriminator overfitting, and leads to further boost on limited-scale datasets. Additionally,
our novel contrastive loss generalizes discriminator representations, makes them more distinguishable between real and fake, and in turn incentivizes better generation quality.
\section*{Acknowledgement}

Ning Yu was partially supported by Twitch Research Fellowship. This work was also partially supported by the DARPA SAIL-ON (W911NF2020009) program. Any opinions, findings, conclusions, or recommendations expressed in this material are those of the authors and do not necessarily reflect the views of the DARPA. We thank Tero Karras, Xun Huang, and Tobias Ritschel for constructive advice.

{\small
\bibliographystyle{ieee_fullname}
\bibliography{main}
}

\clearpage
\renewcommand\thesubsection{\Alph{subsection}}
\section{Supplementary material}

\subsection{Different GAN backbones for dual contrastive loss}
In Table~\ref{tab:comparisions_different_backbones_loss} we show the consistent and significant advantages of our dual contrastive loss on two other GAN backbones: SNGAN~\cite{miyato2018spectral} and StyleGAN~\cite{karras2019StyleGAN}.

\subsection{Self-attention at different generator resolutions}
It is empirically acknowledged that the optimal resolution to replace convolution with self-attention in the generator is specific to dataset and image resolution~\cite{zhang2019self}. For the state-of-the-art attention module SAN~\cite{zhao2020exploring} in Table~\ref{tab:attn_ablation} in the main paper, we find that it achieves the optimal performance at 32$\times$32 generator resolution consistently over all the limited-scale 128$\times$128 datasets, and therefore we report these FIDs.

For the large-scale datasets with varying resolutions in Table~\ref{tab:main_results} in the main paper, we conduct an analysis study on their optimal resolutions as shown in Table~\ref{supp_tab:analysis_G_res}.

We find there is a specific optimal resolution for each dataset, and the FID turns monotonically deteriorated when introducing self-attention one resolution up or down. We reason that each dataset has its own spatial scale and complexity. If longer-range dependency or consistency counts more than local details in one dataset, e.g., CLEVR, it is more favorable to use self-attention in an earlier layer, thus at a lower resolution. We stick to the optimal resolution and report the corresponding FID for each dataset in Table~\ref{tab:main_results} in the main paper.

\subsection{Different reference-attention configurations}
Eq.~\ref{eq:output_ref_attn} in the main paper provides the flexibility of how to cooperate between reference and primary images. We empirically explore the other configurations of sources to the key, query, and value components in the reference-attention. The following two equations, Eq. \ref{supp_eq:output_ref_attn_baseline1} and Eq. \ref{supp_eq:output_ref_attn_baseline2}, correspond to the two configuration variants we compare to.
\begin{equation}
\mathbf{O}^\textit{ref}\doteq\mathsf{attn}\left(\mathbf{K}(\mathbf{T}_\textit{pri}),\mathbf{Q}(\mathbf{T}_\textit{ref}),\mathbf{V}(\mathbf{T}_\textit{ref})\right)+\mathbf{T}_\textit{pri}
\label{supp_eq:output_ref_attn_baseline1}
\end{equation}
\begin{equation}
\mathbf{O}^\textit{ref}\doteq\mathsf{attn}\left(\mathbf{K}(\mathbf{T}_\textit{pri}),\mathbf{Q}(\mathbf{T}_\textit{ref}),\mathbf{V}(\mathbf{T}_\textit{pri})\right)+\mathbf{T}_\textit{pri}
\label{supp_eq:output_ref_attn_baseline2}
\end{equation}

From Table~\ref{supp_tab:analysis_D_config}, we validate that Eq.~\ref{eq:output_ref_attn} in the main paper is the best setting. We reason that the value embedding is relatively independent of the key and query embeddings. Hence we should encode value from one source, and key and query from the other source. Also, because the value and residual shortcut contribute more directly to the discriminator output, we should feed them with the primary image, and feed the key and query with the reference image to formulate the spatially adaptive kernel.

\begin{table}[t!]
\begin{center}
\small
\begin{tabular}{l|ccccc}
\toprule
Method & FFHQ & Bedroom & Church & Horse & CLEVR \\
\midrule
SNGAN & 11.28 & 11.14 & 7.37 & 13.87 & 29.19 \\
+ Contr & \textbf{8.98} & \textbf{10.79} & \textbf{6.51} & \textbf{13.59} & \textbf{18.23} \\
\midrule
StyleGAN & 6.83 & 5.30 & 5.12 & 7.27 & 12.43 \\
+ Contr & \textbf{6.42} & \textbf{4.76} & \textbf{4.48} & \textbf{6.26} & \textbf{8.96} \\
\bottomrule
\end{tabular}
\end{center}
\caption{Comparisons in FID on different GAN backbones.}
\label{tab:comparisions_different_backbones_loss}
\end{table}

\begin{table}[t!]
\begin{center}
\small
\begin{tabular}{c|ccccc}
\toprule
Resolution & FFHQ & Bedroom & Church & Horse & CLEVR \\
\midrule
$8^2$ & 6.08 & 4.43 & 5.10 & 4.24 & 10.44 \\
$16^2$ & 5.81 & 4.21 & 5.24 & \textbf{3.58} & \textbf{8.96} \\
$32^2$ & 5.69 & \textbf{3.48} & \textbf{4.38} & 3.75 & 9.04 \\
$64^2$ & \textbf{5.13} & 3.69 & 4.57 & 3.94 & 12.48 \\
$128^2$ & 5.75 & 6.69 & 4.82 & 6.82 & 18.40 \\
\bottomrule
\end{tabular}
\end{center}
\caption{FID w.r.t. the resolution at which we replace convolution with SAN~\cite{zhao2020exploring} in the generator.}
\label{supp_tab:analysis_G_res}
\end{table}

\begin{table}[t!]
\begin{center}
\small
\resizebox{\linewidth}{!}{
\begin{tabular}{c|cccc}
\toprule
Configuration & CelebA & Animal Face  & Bedroom  & Church  \\
\midrule
Eq.~\ref{supp_eq:output_ref_attn_baseline1} &  10.39 & 65.16 & 20.22 & 17.85 \\
Eq.~\ref{supp_eq:output_ref_attn_baseline2} & 10.95 & 32.33 & 11.05 & 8.33 \\
Eq.~\ref{eq:output_ref_attn} in main & \textbf{7.48} & \textbf{31.08} & \textbf{8.32}& \textbf{7.86} \\
\bottomrule
\end{tabular}}
\end{center}
\caption{FID w.r.t. different reference-attention configurations in the discriminator. For computationally efficient comparisons, we use the 30k subset of each dataset at 128$\times$128 resolution.}
\label{supp_tab:analysis_D_config}
\end{table}

\begin{table}[t!]
\begin{center}
\small
\resizebox{\linewidth}{!}{
\begin{tabular}{c|cccc}
\toprule
Resolution & CelebA & Animal Face  & Bedroom  & Church  \\
\midrule
$8^2$ & \textbf{7.48} & \textbf{31.08} & \textbf{8.32}& \textbf{7.86} \\
$16^2$ &  31.36 & 118.82 & 11.05 & 11.42 \\
$32^2$ & 55.07 & 195.82 & 146.85 & 61.83 \\
\bottomrule
\end{tabular}}
\end{center}
\caption{FID w.r.t. the resolution at which we replace convolution with reference-attention in the discriminator. For computationally efficient comparisons, we use the 30k subset of each dataset at 128$\times$128 resolution.}
\label{supp_tab:analysis_D_res}
\end{table}

\subsection{Reference-attention at different discriminator resolutions}
In Table~\ref{supp_tab:analysis_D_res}, we analyze the relationship between generation quality and the resolution to replace convolution with reference-attention in the discriminator. We stop investigation to higher resolutions because the training turns easily diverging. We conclude introducing reference-attention at the lowest possible resolution is most beneficial. We reason that the deepest features are the most representative for cooperating between reference and primary images. Also because the primary and reference images are not pre-aligned, the lowest resolution covers the largest receptive field and therefore leads to the largest overlap between the two images that should be corresponded. We stick to the 8$\times$8 resolution for all the experiments involving reference-attention.

\begin{table*}[t!]
\begin{center}
\small
\begin{tabular}{c|cc|cc|cc|cc}
\toprule
& \multicolumn{2}{c}{CelebA} & \multicolumn{2}{c}{Animal Face} & \multicolumn{2}{c}{Bedroom} & \multicolumn{2}{c}{Church} \\
Data size & StyleGAN2 & + ref attn & StyleGAN2 & + ref attn & StyleGAN2 & + ref attn & StyleGAN2 & + ref attn \\
\midrule
1K & 55.71 & \textbf{43.19} & 181.26 & \textbf{123.08} & 230.40 & \textbf{79.81} & 107.31 & \textbf{43.05}\\
5K & 23.48 & \textbf{18.48} & 89.88 & \textbf{61.17} & 57.68 & \textbf{19.64} & 29.30 & \textbf{17.85}\\
10K & 14.73 & \textbf{12.72} & 61.36 & \textbf{45.49} & 40.70 & \textbf{12.29} & 17.94 & \textbf{12.13}\\
30K & 9.84 & \textbf{7.48} & 36.55 & \textbf{31.08} & 19.33 & \textbf{8.32} & 11.02 & \textbf{7.86} \\
50K & \textbf{6.59} & 7.09 & 28.92 & \textbf{28.43} & 14.01 & \textbf{7.15} & 8.88 & \textbf{7.09}\\
100K & \textbf{5.61} & 6.86 & \textbf{22.85} & 28.37 & 9.42 & \textbf{6.89} & 7.32 & \textbf{7.08}\\
\bottomrule
\end{tabular}
\end{center}
\caption{Comparisons in FID between StyleGAN2 config E baseline and that with our reference-attention in the discriminator. Our method consistently improves the baseline when dataset size varies between 1k and 30k images. For computationally efficient comparisons, we use each dataset at 128$\times$128 resolution. See Fig.~\ref{fig:comparisons_dataset_size} in the main paper for the corresponding plots.}
\label{supp_tab:comparisons_dataset_size_ref_attn}
\end{table*}

\begin{table*}[t!]
\begin{center}
\resizebox{\linewidth}{!}{%
\begin{tabular}{ll|ccccc|cccc|cccc|cccc|cccc}
\toprule
& & \multicolumn{5}{c}{FFHQ} & \multicolumn{4}{c}{Bedroom} & \multicolumn{4}{c}{Church} & \multicolumn{4}{c}{Horse} & \multicolumn{4}{c}{CLEVR} \\
Method & Loss & FID & PPL & P & R & Sep & FID & PPL & P & R & FID & PPL & P & R & FID & PPL & P & R & FID & PPL & P & R\\
\midrule
BigGAN~\cite{brock2018BigGAN} &  Adv & 11.4 & - & - & - & - & - & - & - & - & - & - & - & - & - & - & - & - & - & - & - & - \\
U-Net GAN~\cite{schonfeld2020u} & Adv & 7.48 & \textbf{32} & 0.68 & 0.19 & \textbf{2.00} & 17.6 & \textbf{504} & 0.48 & 0.03 & 11.7 & \textbf{318} & \textbf{0.62} & 0.07 & 20.2 & \textbf{296} & 0.57 & 0.13 & 33.3 & 202 & 0.04 & 0.08 \\
StyleGAN2~\cite{karras2019StyleGAN2} & Adv & 4.86 & \underline{47} & 0.69 & \underline{0.42} & 5.08 & 4.01 & \underline{976} & \underline{0.59} & 0.32 & 4.54 & \underline{511} & 0.57 & \underline{0.42} & 3.91 & 637 & 0.63 & \underline{0.40} & 9.62 & 582 & 0.46 & 0.56 \\
\midrule
StyleGAN2 w/ attn & Adv & 5.13 & 54 & 0.69 & 0.41 & 4.18 & \underline{3.48} & 1384 & \underline{0.59} & \underline{0.36} & 4.38 & 611 & 0.59 & 0.41 & \underline{3.59} & \underline{636} & \textbf{0.64} & 0.39 & 8.96 & \textbf{67} & 0.47 & 0.63 \\
StyleGAN2& Contr  & \textbf{3.98} & 50 & \textbf{0.71} & \textbf{0.44} & 3.76 & 3.86 & 1054 & \textbf{0.60} & 0.31 & \underline{3.73} & 619 & \underline{0.60} & 0.40 & 3.70 & 740 & \textbf{0.64} & 0.39 & \underline{6.06} & 816 & \underline{0.57} & \underline{0.65} \\
StyleGAN2 w/ attn & Contr & \underline{4.63} & 65 & \underline{0.70} & 0.41 & \underline{3.60} & \textbf{3.31} & 1830 & \underline{0.59} & \textbf{0.37} & \textbf{3.39} & 1239 & \underline{0.60} & \textbf{0.45} & \textbf{2.97} & 1367 & \textbf{0.64} & \textbf{0.43} & \textbf{5.05} & \underline{106} & \textbf{0.58} & \textbf{0.70} \\
\bottomrule
\end{tabular}}
\end{center}
\caption{Comparisons to the state-of-the-art GANs in various metrics on the large-scale datasets. We highlight the best in \textbf{bold} and second best with \underline{underline}. ``w/ attn'' indicates using the self-attention in the generator. ``Contr'' indicates using our dual contrastive loss instead of conventional GAN loss.}
\label{supp_tab:main_results_various_metrics}
\end{table*}

\begin{table*}[t!]
\begin{center}
\small
\begin{tabular}{ll|ccccc}
\toprule
Method & Loss & CelebA & Animal Face & Bedroom & Church \\
\midrule
StyleGAN2~\cite{karras2019StyleGAN2} & Adv & 9.84 & 36.55 & 19.33 & 11.02 \\
StyleGAN2 w/ self-attn-G & Adv & 8.60 & 32.72 & 16.36 & 9.62 \\
StyleGAN2 w/ self-attn-G & Contr & 7.55 & 25.83 & 10.99 & 8.12 \\
StyleGAN2 w/ self-attn-G ref-attn-D & Adv & 7.48 & 31.08 & \textbf{8.32} & \textbf{7.86} \\
StyleGAN2 w/ self-attn-G ref-attn-D & Contr & \textbf{6.00} & \textbf{25.03} & 12.84 & 8.75 \\
\bottomrule
\end{tabular}
\end{center}
\caption{Comparisons in FID to StyleGAN2 config E baseline on the limited-scale datasets. Our configurations consistently improve the baseline, the relative improvements of which are even more significant than those on the large-scale datasets. We use the 30k subset of each dataset at 128$\times$128 resolution.}
\label{supp_tab:comparisons_limited_data}
\end{table*}

\subsection{FID w.r.t. data size for reference-attention}
We report in Table~\ref{supp_tab:comparisons_dataset_size_ref_attn} the detailed values from Fig.~\ref{fig:comparisons_dataset_size} in the main paper. Our method consistently improves the baseline when dataset size varies between 1k and 30k images.

\subsection{Comparisons to the state of the art in various metrics}
We extend Table~\ref{tab:main_results} in the main paper with additional evaluation metrics for GANs, which are proposed and used in StyleGAN~\cite{karras2019StyleGAN} and/or StyleGAN2~\cite{karras2019StyleGAN2}: Perceptual Path Length (PPL), Precision (P),  Recall (R), and Separability (Sep). See Table~\ref{supp_tab:main_results_various_metrics}.

Consistent with FID rankings, our attention modules and dual contrastive loss also improve from StyleGAN2 baseline for Precision, Recall, and Separability in most cases. It is worth noting that the rankings of PPL are negatively correlated to all the other metrics, which disqualifies it as an effective evaluation metric in our experiments. E.g., U-Net GAN has the best PPL in most cases but in fact it contradicts against its worst FID and worst visual quality in Fig.~\ref{supp_fig:samples_FFHQ}, \ref{supp_fig:samples_Bedroom}, \ref{supp_fig:samples_Church}, \ref{supp_fig:samples_Horse}, and \ref{supp_fig:samples_CLEVR}.

\subsection{Comparisons on the limited-scale datasets}
Besides comparisons on the large-scale datasets, we also compare to StyleGAN2~\cite{karras2019StyleGAN2} baseline on the limited-scale datasets in Table~\ref{supp_tab:comparisons_limited_data}. We use the 30k subset of each dataset at 128$\times$128 resolution. We find:

(1) Comparing across the first, second, and third rows, self-attention generator, dual contrastive loss, and their synergy significantly and consistently improve on all the limited-scale datasets, more than what they improve on the large-scale datasets: from 18.1\% to 23.3\% on CelebA~\cite{liu2015deep} and Animal Face~\cite{liu2019few}, from 17.5\% to 43.2\% on LSUN Bedroom~\cite{yu15lsun}, and from 25.2\% to 26.4\% on LSUN Church~\cite{yu15lsun}. It indicates the limited-scale setting is more challenging and leaves more space for our improvements.

(2) Comparing between the first and fourth rows, the reference-attention discriminator improves significantly and consistently on all the datasets up to 57.0\% on LSUN Bedroom. We reason that the arbitrary pair-up between reference and primary images results in a beneficial effect similar in spirit to data augmentation, and consequently generalizes the discriminator representation and mitigates its overfitting.

(3) However, according to the fifth row, reference-attention discriminator is sometimes not compatible with contrastive learning because they may together overly augment the classification task: contrastive learning for one pair of primary and reference input against a batch of other pairs makes adversarial training unstable. This observation differs from that of pairwise contrastive learning in the unsupervised learning scenario~\cite{hadsell2006dimensionality,tschannen2019mutual,chen2020simple,chen2020big} or GAN applications with reconstructive regularization~\cite{park2020contrastive}.

Even though in this paper our main scope is GANs on large-scale datasets, we believe these findings to be very interesting for researchers to design their networks for limited-scale datasets.

\subsection{Uncurated generated samples}
\begin{figure*}
\centering
\includegraphics[width=\linewidth]{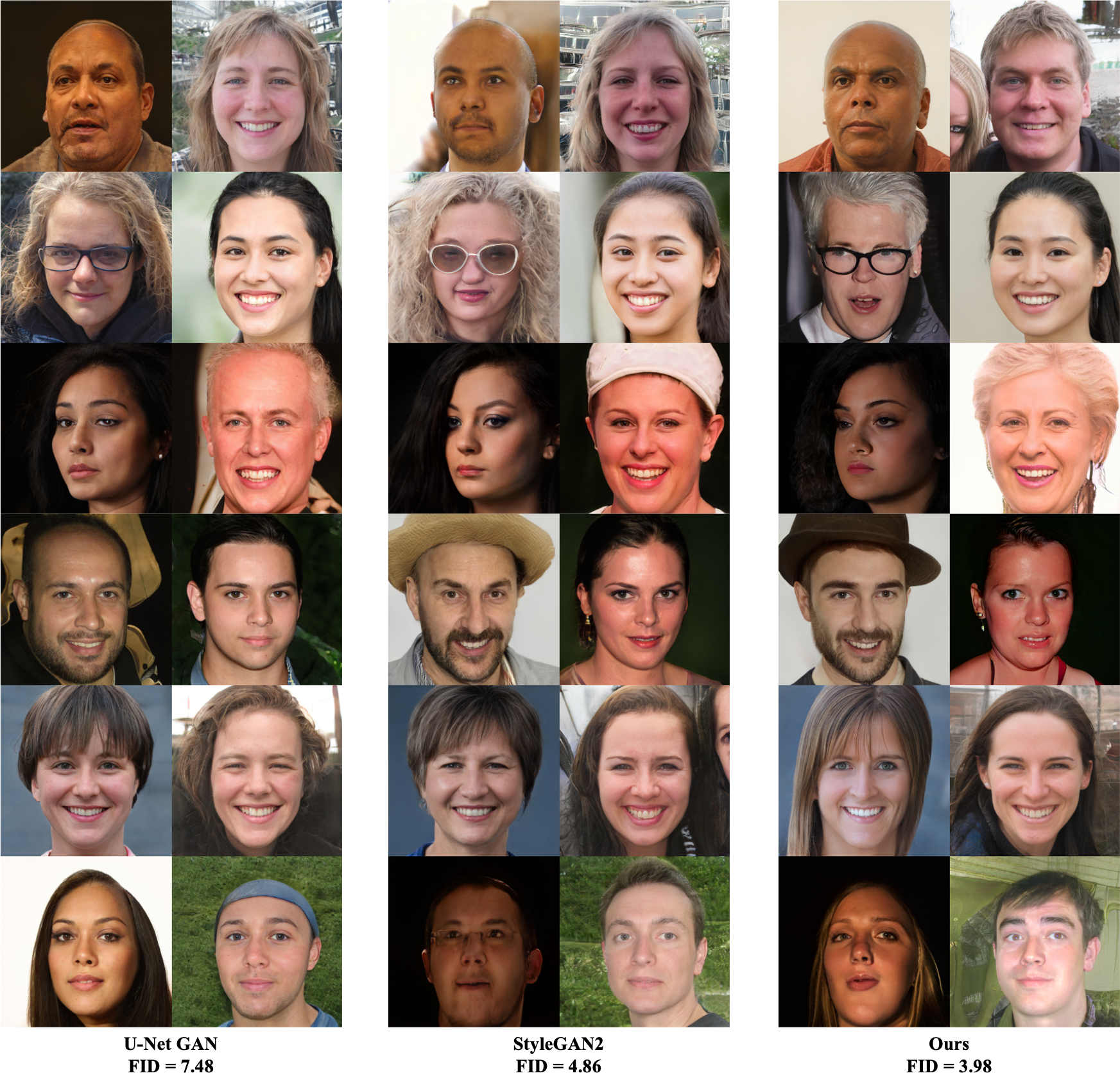}
\caption{Uncurated generated samples at 256$\times$256 for FFHQ dataset~\cite{karras2019StyleGAN}. To align the comparisons, we use the same real query images for pre-trained generators to reconstruct. Our generation significantly outperforms the baselines in terms of quality, long-range dependencies, and spatial consistency.}
\label{supp_fig:samples_FFHQ}
\end{figure*}

\begin{figure*}
\centering
\includegraphics[width=\linewidth]{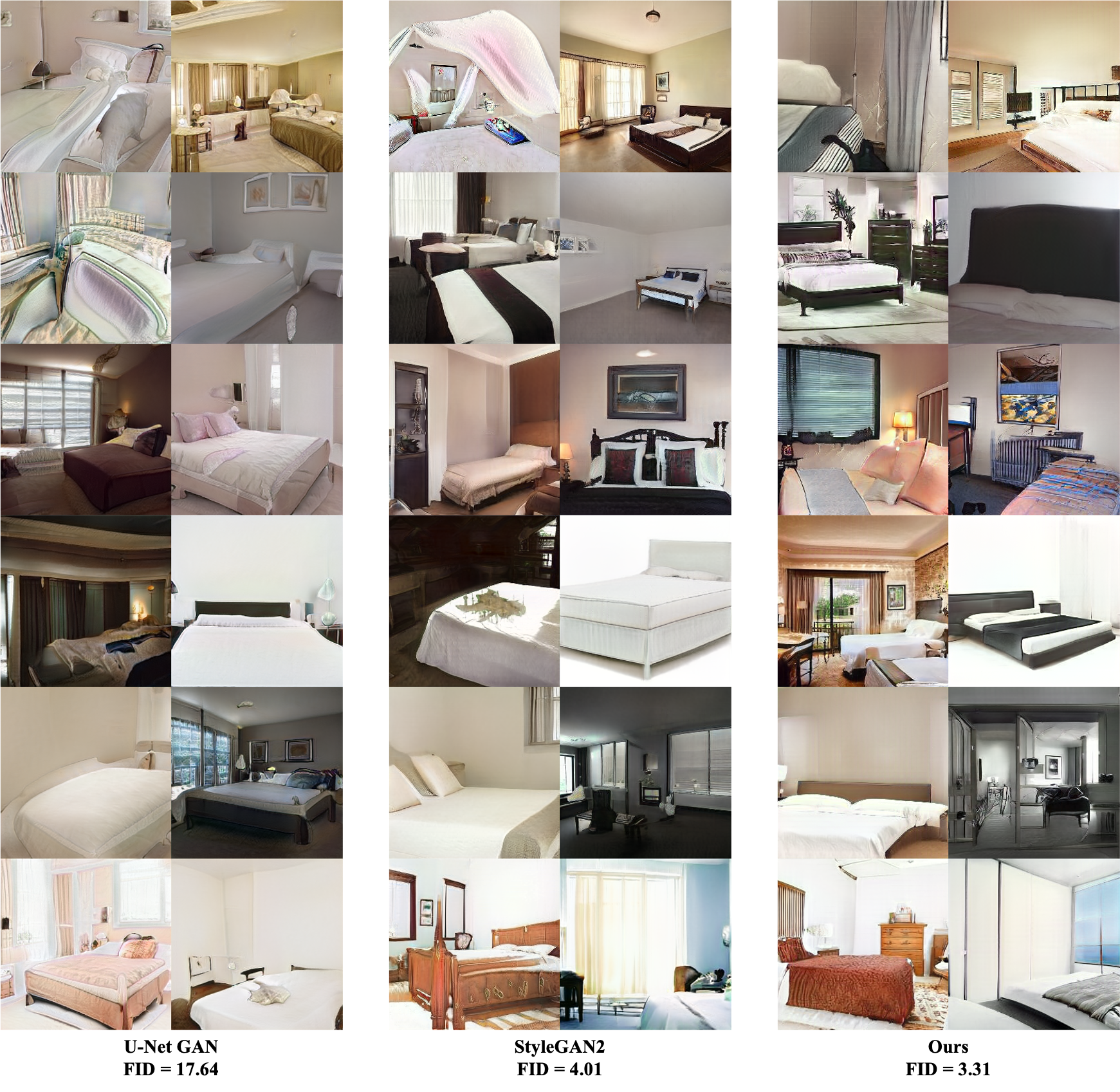}
\caption{Uncurated generated samples at 256$\times$256 for LSUN Bedroom dataset~\cite{yu15lsun}. To align the comparisons, we use the same real query images for pre-trained generators to reconstruct. Our generation significantly outperforms the baselines in terms of quality, long-range dependencies, and spatial consistency.}
\label{supp_fig:samples_Bedroom}
\end{figure*}

\begin{figure*}
\centering
\includegraphics[width=\linewidth]{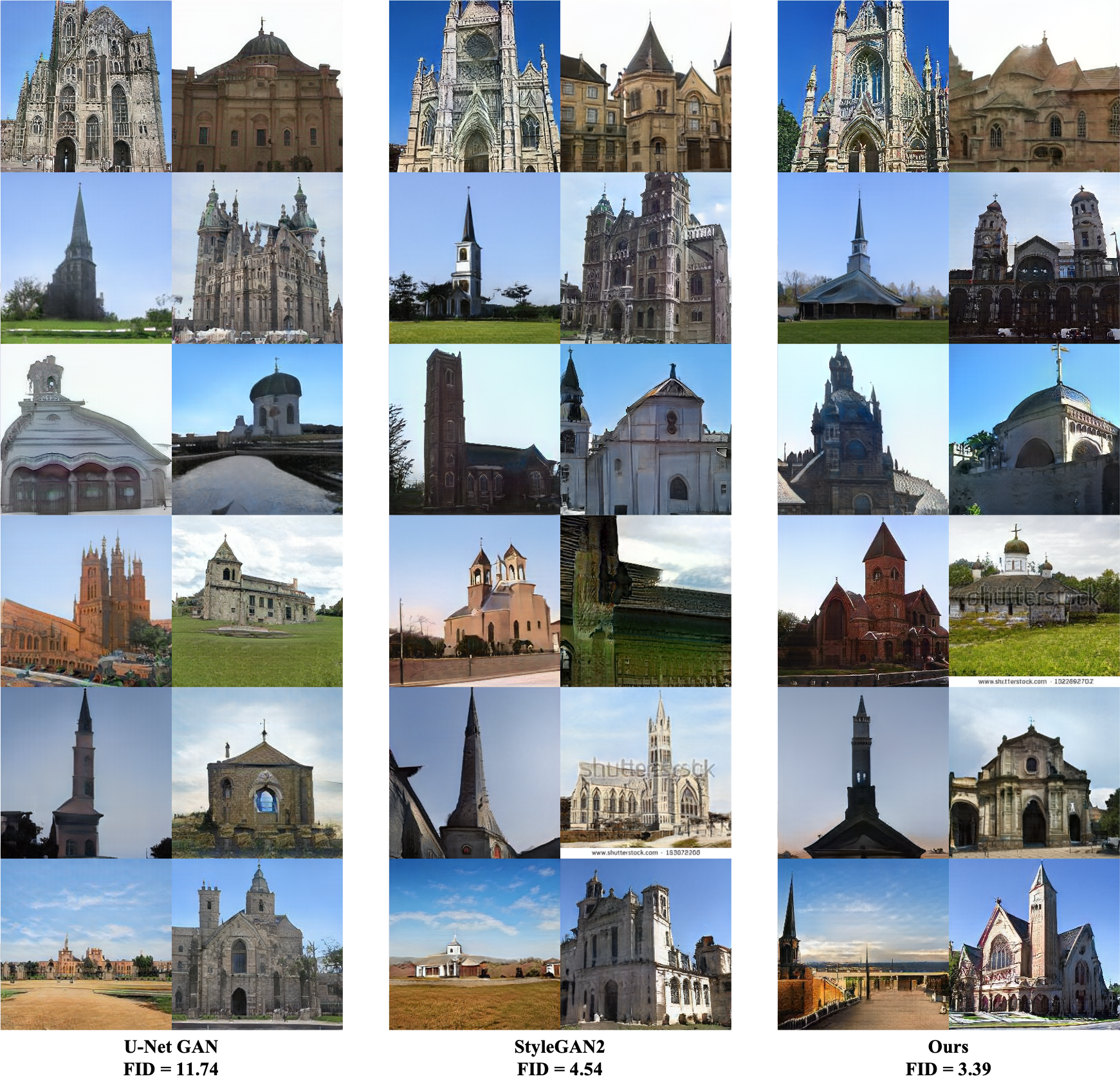}
\caption{Uncurated generated samples at 256$\times$256 for LSUN Church dataset~\cite{yu15lsun}. To align the comparisons, we use the same real query images for pre-trained generators to reconstruct. Our generation significantly outperforms the baselines in terms of quality, long-range dependencies, and spatial consistency.}
\label{supp_fig:samples_Church}
\end{figure*}

\begin{figure*}
\centering
\includegraphics[width=\linewidth]{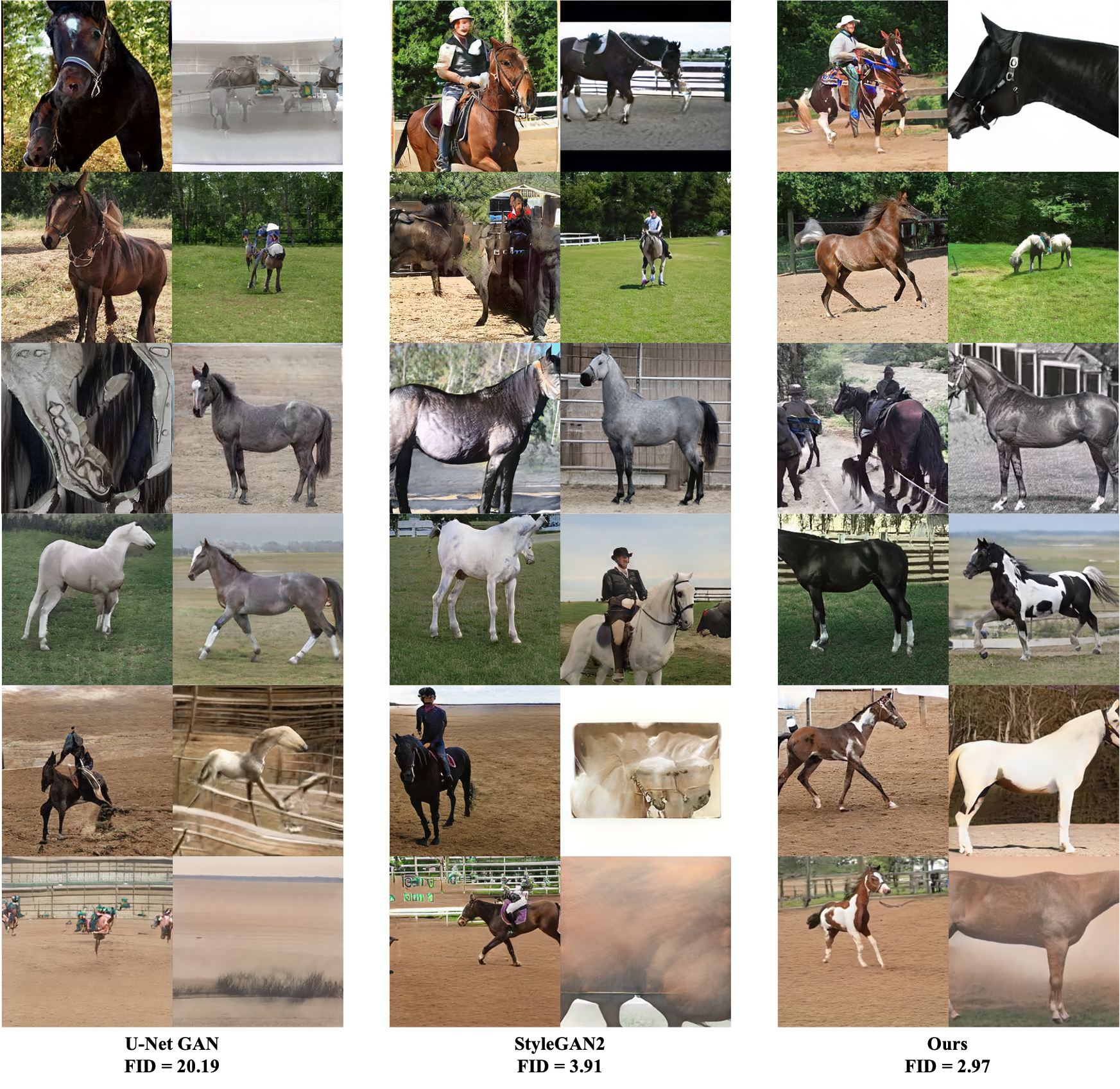}
\caption{Uncurated generated samples at 256$\times$256 for LSUN Horse dataset~\cite{yu15lsun}. To align the comparisons, we use the same real query images for pre-trained generators to reconstruct. Our generation significantly outperforms the baselines in terms of quality, long-range dependencies, and spatial consistency.}
\label{supp_fig:samples_Horse}
\end{figure*}

\begin{figure*}
\centering
\includegraphics[width=\linewidth]{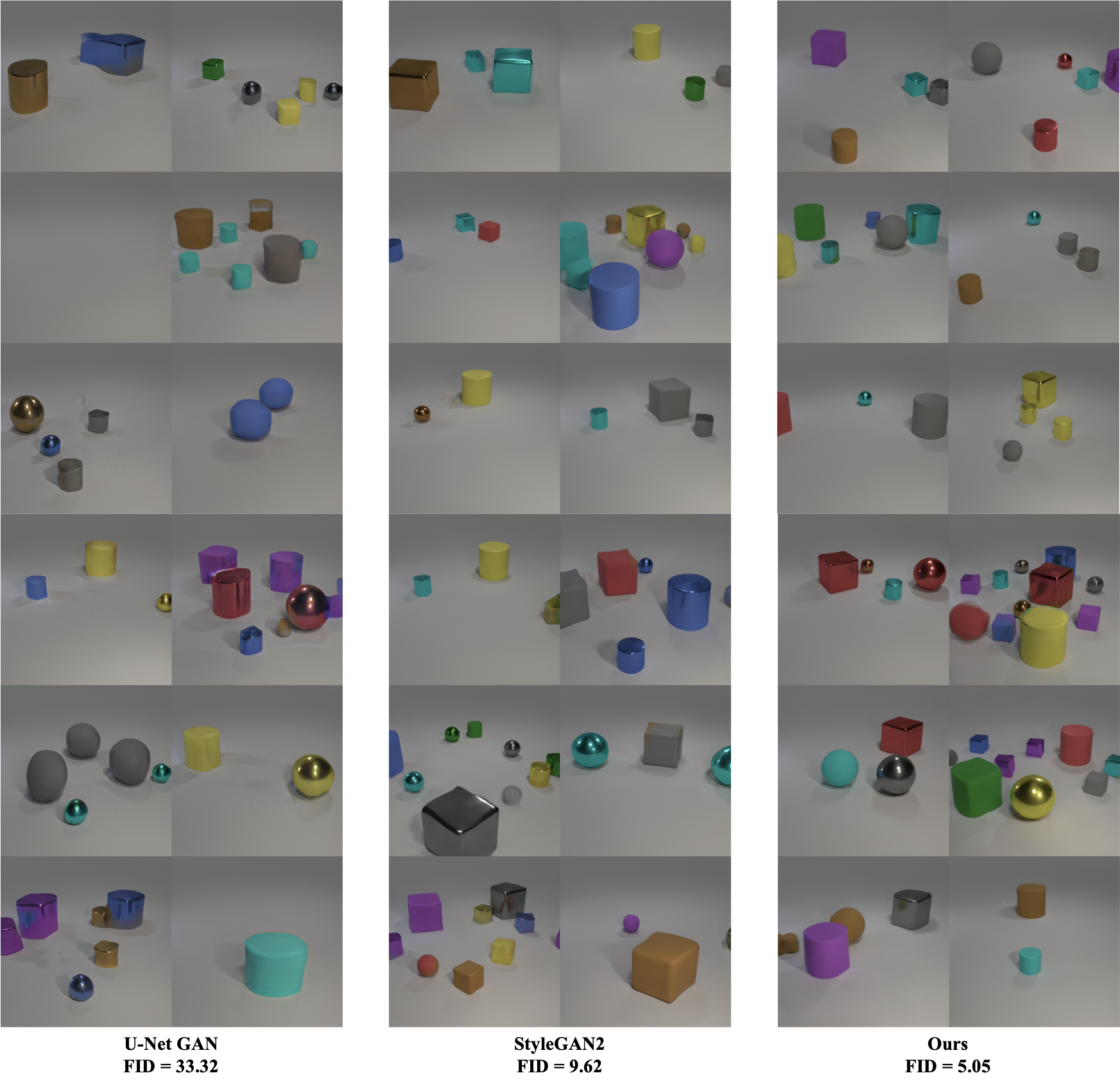}
\caption{Uncurated generated samples at 256$\times$256 for CLEVR dataset~\cite{johnson2017clevr}. To align the comparisons, we use the same real query images for pre-trained generators to reconstruct. Our generation significantly outperforms the baselines in terms of quality, long-range dependencies, and spatial consistency.}
\label{supp_fig:samples_CLEVR}
\end{figure*}

For comparisons to the state of the art, we show more uncurated generated samples in Figure~\ref{supp_fig:samples_FFHQ}, \ref{supp_fig:samples_Bedroom}, \ref{supp_fig:samples_Church}, \ref{supp_fig:samples_Horse} and \ref{supp_fig:samples_CLEVR}. Our generation significantly outperforms the baselines U-Net GAN~\cite{schonfeld2020u} and StyleGAN2~\cite{karras2019StyleGAN2} in terms of quality, long-range dependencies, and spatial consistency.

\subsection{Self-attention maps}
\begin{figure*}
\centering
\includegraphics[width=\linewidth]{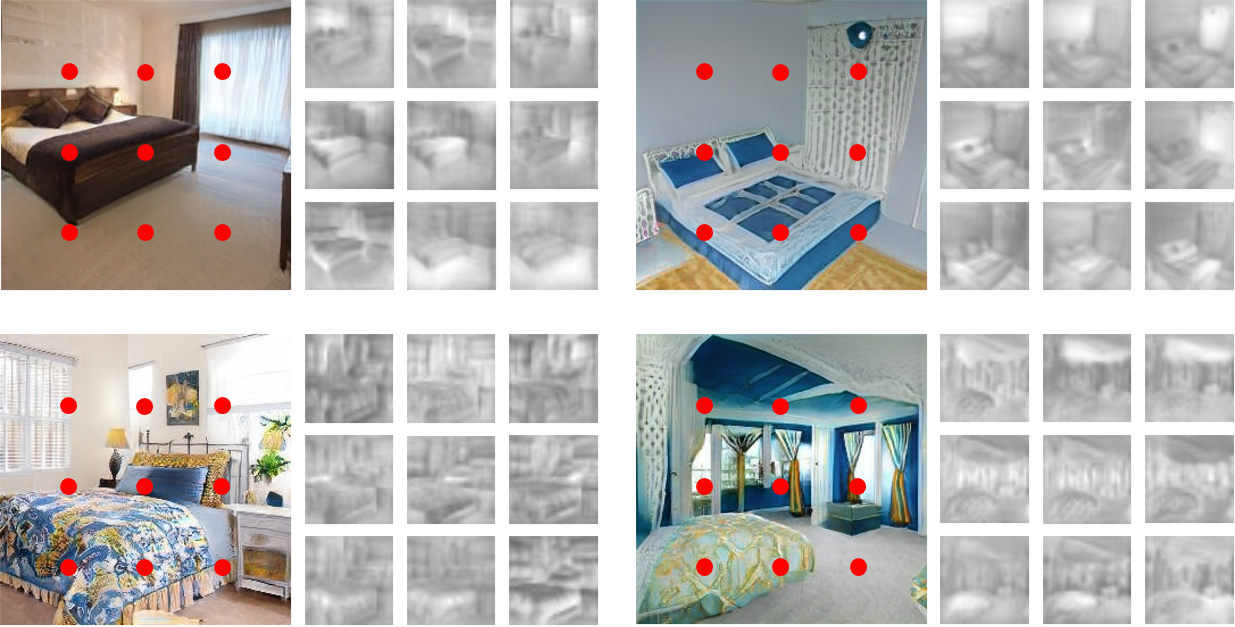}
\caption{StyleGAN2 + SAN generated LSUN Bedroom~\cite{yu15lsun} samples at 256$\times$256 and their self-attention maps at 32$\times$32 in the generator for the corresponding dot positions. Considering there is an attention weight kernel $\mathbf{w}\in\mathbb{R}^{s \times s \times c}$ for each position, we visualize the norm for each spatial position of $\mathbf{w}$. The attention maps strongly align to the semantic layout and structures of the generated images, which enable long-range dependencies across objects.}
\label{supp_fig:attention_Bedroom}
\end{figure*}

\begin{figure*}
\centering
\includegraphics[width=\linewidth]{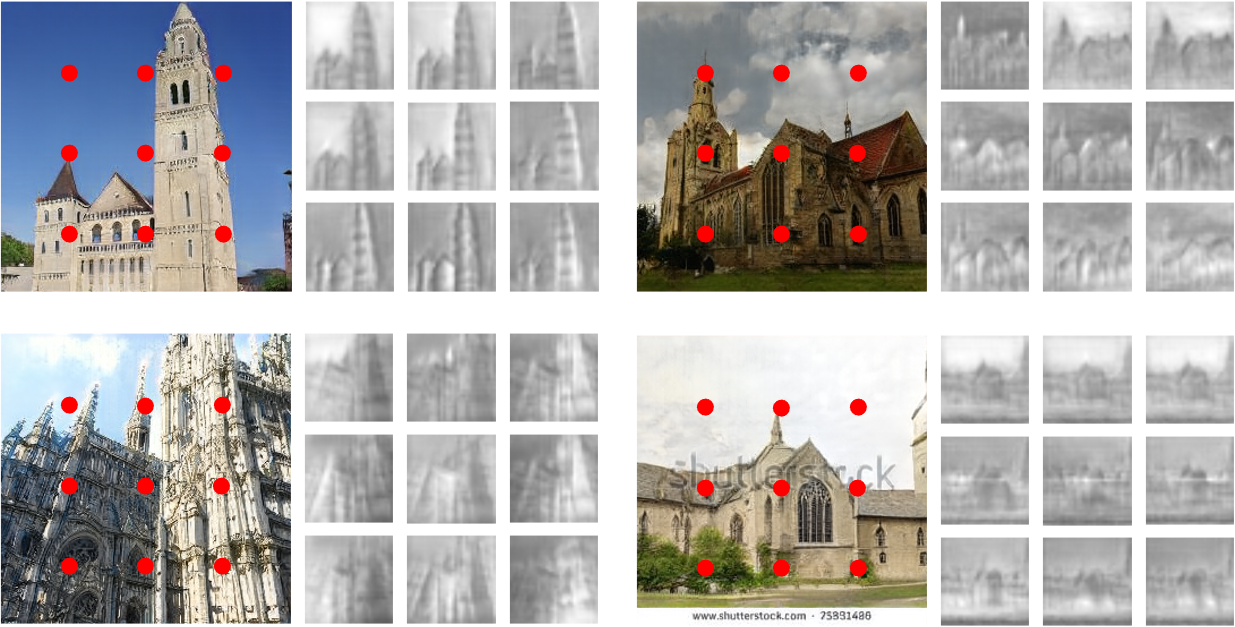}
\caption{StyleGAN2 + SAN generated LSUN Church~\cite{yu15lsun} samples at 256$\times$256 and their self-attention maps at 32$\times$32 in the generator for the corresponding dot positions. Considering there is an attention weight kernel $\mathbf{w}\in\mathbb{R}^{s \times s \times c}$ for each position, we visualize the norm for each spatial position of $\mathbf{w}$. The attention maps strongly align to the semantic layout and structures of the generated images, which enable long-range dependencies across objects.}
\label{supp_fig:attention_Church}
\end{figure*}

\begin{figure*}
\centering
\includegraphics[width=\linewidth]{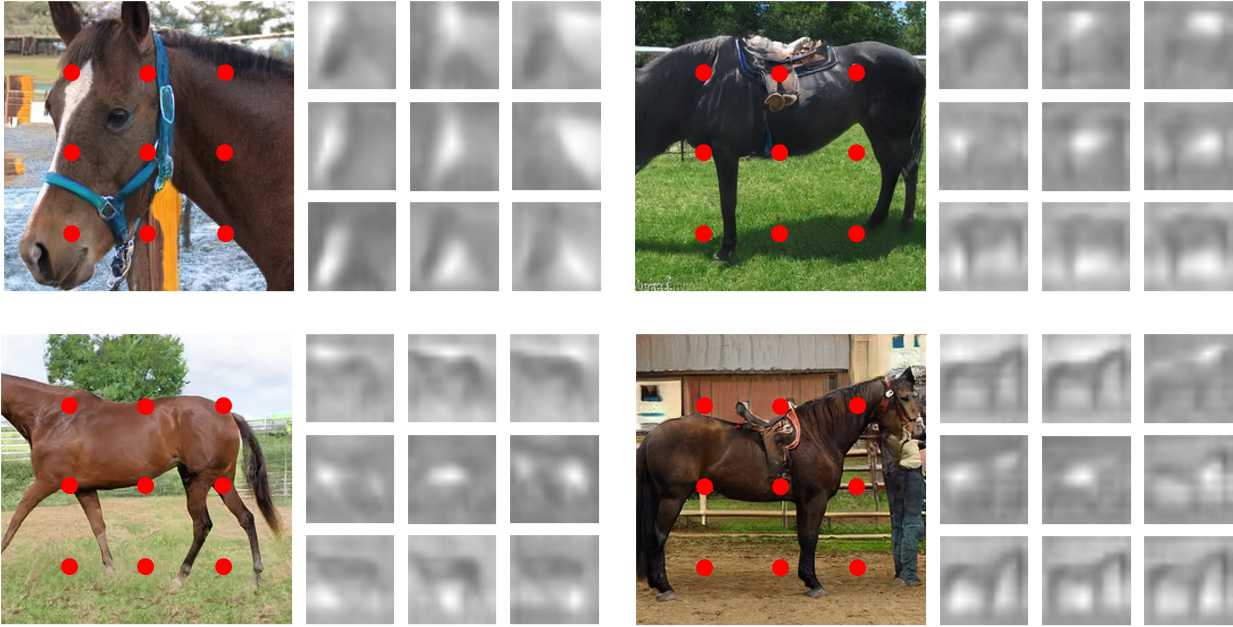}
\caption{StyleGAN2 + SAN generated LSUN Horse~\cite{yu15lsun} samples at 256$\times$256 and their self-attention maps at 16$\times$16 in the generator for the corresponding dot positions. Considering there is an attention weight kernel $\mathbf{w}\in\mathbb{R}^{s \times s \times c}$ for each position, we visualize the norm for each spatial position of $\mathbf{w}$. The attention maps strongly align to the semantic layout and structures of the generated images, which enable long-range dependencies across objects.}
\label{supp_fig:attention_Horse}
\end{figure*}

\begin{figure*}
\centering
\includegraphics[width=\linewidth]{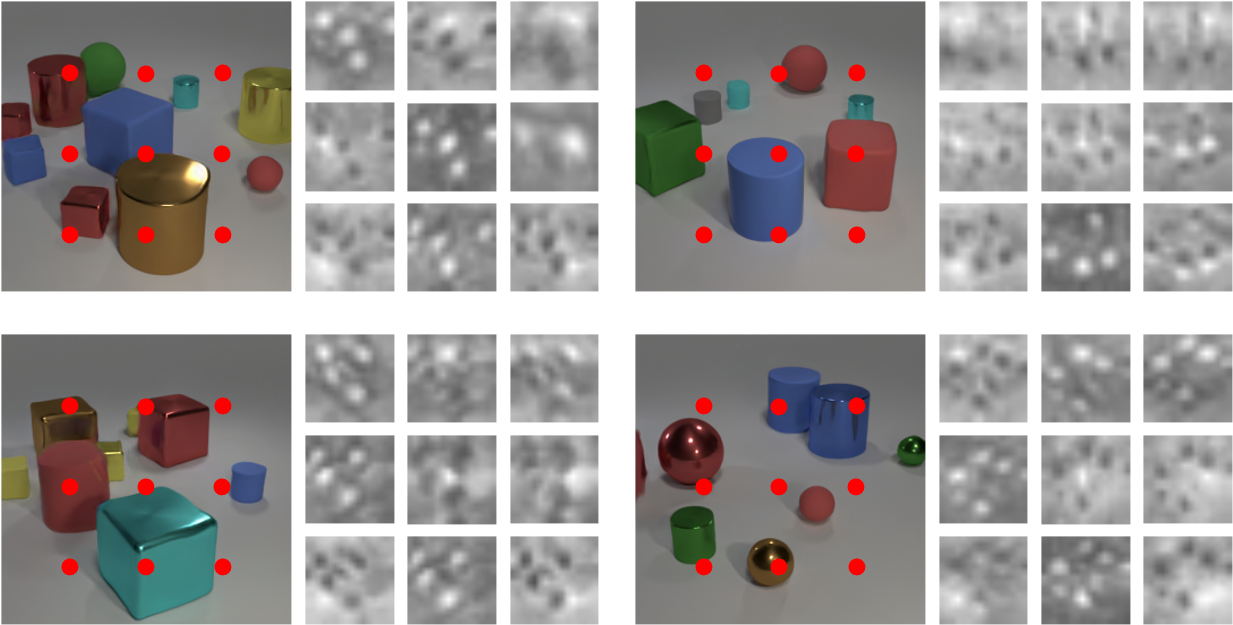}
\caption{StyleGAN2 + SAN generated CLEVR~\cite{johnson2017clevr} samples at 256$\times$256 and their self-attention maps at 16$\times$16 in the generator for the corresponding dot positions. Considering there is an attention weight kernel $\mathbf{w}\in\mathbb{R}^{s \times s \times c}$ for each position, we visualize the norm for each spatial position of $\mathbf{w}$. The attention maps strongly align to the semantic layout and structures of the generated images, which enable long-range dependencies across objects.}
\label{supp_fig:attention_CLEVR}
\end{figure*}

For self-attention maps in the generator, we show more results in Figure~\ref{supp_fig:attention_Bedroom}, \ref{supp_fig:attention_Church}, \ref{supp_fig:attention_Horse}, and \ref{supp_fig:attention_CLEVR}. The attention maps strongly align to the semantic layout and structures of the generated images, which enable long-range dependencies across objects.

\end{document}